%% file: main.tex
\pdfoutput=1
\PassOptionsToPackage{table}{xcolor,dvipsnames}

\documentclass[11pt]{article}

\usepackage[final]{acl}

\usepackage{times}
\usepackage{latexsym}

\usepackage[T1]{fontenc}
\usepackage[utf8]{inputenc}

\usepackage{microtype}

\usepackage{inconsolata}

\usepackage{graphicx}

\usepackage{hyperref}
\usepackage{url}

\usepackage{booktabs}
\usepackage{wrapfig}
\usepackage{adjustbox}
\usepackage{multirow}
\usepackage{caption}
\usepackage{subcaption}
\usepackage{amssymb}
\usepackage{amsmath}
\usepackage{tikz}
\usepackage{colortbl}
\usepackage{rotating}

\usepackage{enumitem}
\usepackage{cleveref}
\usepackage{balance}

\definecolor{darkspringgreen}{rgb}{0.09, 0.45, 0.27}
\definecolor{brickred}{rgb}{0.71373,0.19608,0.1098}

\definecolor{figgreen}{rgb}{0.78,0.89,0.87}
\definecolor{figpink}{rgb}{0.949,0.776,0.87}
\definecolor{figorange}{rgb}{0.968,0.85,0.768}
\definecolor{figyellow}{rgb}{0.98,0.93,0.796}
\definecolor{figpurple}{rgb}{0.858,0.8,0.94}
\definecolor{figblue}{rgb}{0.322,0.671,0.424}
\def\llama{Llama3-8B-Instruct}
\def\qwen{Qwen2-7B-Instruct}
\def\mixtral{Mixtral-8x22B-Instruct}
\def\latin{Latin}
\def\nonlatin{non-Latin}

\def\wiki{Aya-Wiki}
\def\flores{FLORES}
\def\mlqa{Aya-MLQA}

\def\script{\texttt{Script}}
\def\ipa{\texttt{IPA}}
\def\mixed{\texttt{Mixed}}

\def\mixicl{\texttt{Mixed-ICL}}
\def\ipaicl{\texttt{IPA-ICL}}
\def\scripticl{\texttt{Script-ICL}}
\def\romanicl{\texttt{Roman-ICL}}
\def\allicl{\texttt{All-ICL}}
\def\ayamlqa{Aya-MLQA}
\def\orimlqa{Orig. MLQA}

\title{Prompting with Phonemes: Enhancing LLMs' \\ Multilinguality for non-Latin Script Languages}

\author{
\textbf{Hoang Nguyen}$^{1}$\thanks{Work done during an internship at ServiceNow.}\quad
\textbf{Khyati Mahajan}$^{2}$\quad
\textbf{Vikas Yadav}$^{2}$\quad
\textbf{Julian Salazar}$^{3}$
\\
\textbf{Philip S. Yu}$^{1}$\quad
\textbf{Masoud Hashemi}$^{2}$\quad
\textbf{Rishabh Maheshwary}$^{2}$\\
  $^1$University of Illinois at Chicago\quad $^2$ServiceNow\quad
  $^3$Google DeepMind\quad 
\\
\small{\texttt{\{hnguy7,psyu\}@uic.edu}\quad\texttt{julsal@google.com}}\\
\small{\texttt{\{khyati.mahajan,vikas.yadav,masoud.hashemi,rishabh.maheshwary\}@servicenow.com}}\\
}

\begin{document}
\maketitle
\begin{abstract}
Although multilingual LLMs have achieved remarkable performance across benchmarks, we find they continue to underperform on \nonlatin~script languages across contemporary LLM families. This discrepancy arises from the fact that LLMs are pretrained with orthographic scripts, which are dominated by Latin characters that obscure their shared phonology with non-Latin scripts. We propose leveraging phonemic transcriptions as complementary signals to induce script-invariant representations. Our study demonstrates that integrating phonemic signals improves performance across both \nonlatin~and \latin~script languages, with a particularly significant impact on closing the performance gap between the two. Through detailed experiments, we show that phonemic and orthographic scripts retrieve distinct examples for in-context learning (ICL). This motivates our proposed \mixicl~retrieval strategy, where further aggregation from both leads to our significant performance improvements for both \latin~script languages (up to 12.6\%) and \nonlatin~script languages (up to 15.1\%) compared to randomized ICL retrieval.
\end{abstract}

%%%%%%%%%%%%%%%%%%%%%%%%%%%%%%%%%%%%%%%%%%%%%

\section{Introduction} \label{sec:introduction}

\begin{figure}
    \centering
     \includegraphics[trim={5.0cm 5.0cm 0.2cm 2.8cm},clip,width=\columnwidth]{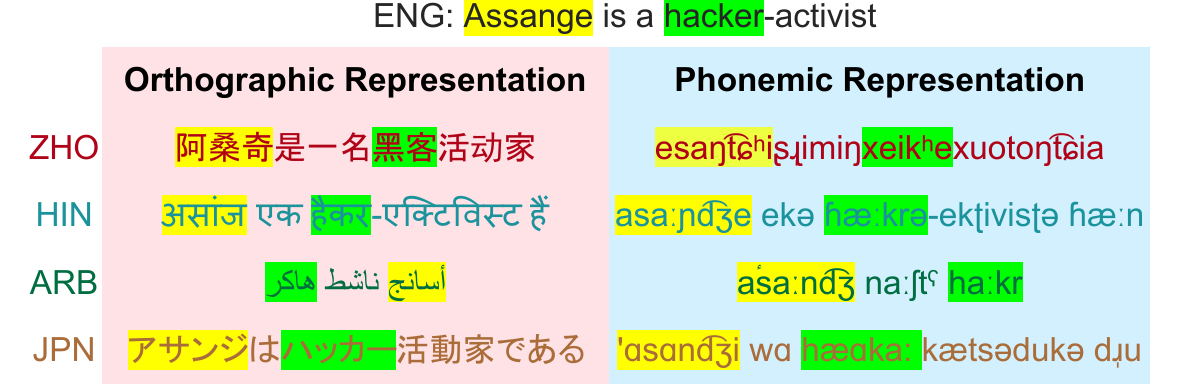}
    \caption{Orthographic and phonemic transcriptions (via the International Phonetic Alphabet, IPA) of the same sentence. Matching colors denote semantically similar words across different languages.  
    }
    \label{fig:example-viz}
\end{figure}

Large language models (LLMs) have demonstrated remarkable multilingual capabilities across various natural language processing (NLP) tasks. The increase in model parameters and rise of instruction datasets have led to the emergent capability of LLMs to perform diverse tasks via few- to zero-shot demonstrations \cite{brown2020language,xia2020cg, wei2022flan,nguyen2023cof} through in-context learning (ICL) during inference \cite{wei2022emergent}. However, these capabilities remain disparate across languages \cite{lai2023chatgpt}, with one particular axis being along \nonlatin~versus \latin~script languages \citep{bang-etal-2023-multitask, ahuja-etal-2023-mega, shliazhko-etal-2024-mgpt}.
To mitigate this disparity, we are motivated by the crucial role of phonemic awareness in human language acquisition and processing, facilitating skills like cross-lingual transfer and reading development \cite{durgunoglu1993cross,spencer2003effects}, in part due to cognates, borrowed words, and shared phonology between language families. We hypothesize that integrating phonemic information could also enable LLMs' robustness to the choice of writing system by capturing such alignments. For instance, in \Cref{fig:example-viz}, the Japanese orthographic representation\footnote{\textit{Orthographic representation} and \textit{textual / written script} are used interchangeably throughout this work.} for \textit{hacker} (\includegraphics[height=1em]{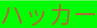}) is significantly different from its English one. However, when observing the phonemic transcriptions---specifically, International Phonetic Alphabet (IPA) transcriptions at the level of phoneme discrimination---one could easily recognize the semantically similar words (\includegraphics[height=1em]{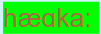}) highlighted in green in \Cref{fig:example-viz}.

On the other hand, current prompting and ICL schemes rely solely on orthographic text input, overlooking potentially valuable linguistic information encoded in the phonemic structure of language. While continual pretraining with phonemic data could enhance LLMs' multilingual capabilities, it faces several challenges. One significant limitation is the scarcity of large-scale multilingual datasets that align orthographic and phonemic transcriptions across diverse languages, especially for less-resourced languages. This lack of aligned data restricts the potential for fine-tuning models on phonemic information at scale. Furthermore, pretraining with phonemic data requires advance planning, data syntheses of vetted quality, and a potential doubling of data size, adding cost and complexity to training.

Hence, we propose integrating phonemic information in LLMs via prompting and via ICL, as these promise a more flexible and resource-efficient approach to realize the integration's benefits. We hypothesize that augmenting with phonemic information could improve both demonstration retrieval and LLM reasoning, by explicitly surfacing fundamental crosslingual information \cite{andersonessentials} that textual scripts might not capture or induce in the LLMs' internal representations. Our contributions include:
\begin{itemize}
    \item Evaluating multilingual performance across contemporary LLM families ($\ge$7B-parameter models) and diverse sets of tasks, with specific focus on Latin vs.\ non-Latin scripts, revealing a significant gap on evaluation metrics (up to 29\%  absolute). We then focus our work on tasks with notable performance disparities such as in text generation (\wiki) and machine translation (\flores).
    \item Investigating the integration of IPA into LLMs via (1) direct prompting (zero- and few-shot) and in (2) retrieval-based ICL augmentation. In particular, we find that simple lexical retrieval ranked using text \text{and} phoneme matching (our proposed \mixicl) gives performance gains of up to 15.1\% relative on generative tasks, together with inference-time gains on \latin~languages. Qualitative analyses examine retrieved cases and validate the observed empirical performance gains.
    \item Analyzing the components involved in phonemic integration and offering insights and guidance for future works aiming to improve LLMs beyond inference-time interventions.
\end{itemize}

%%%%%%%%%%%%%%%%%%%%%%%%%%%%%%%%%%%%%%%%%%%%%

\section{Background and Related Work} \label{sec:related-work}

Phonemes are considered the smallest units of speech distinguishing one word (or word element) from another in a given language.\footnote{\url{https://www.britannica.com/topic/phoneme}} LLMs train on enough text that one may suspect they implicitly learn about the underlying phonemics of text; however, investigative research on the phonemic awareness in language modeling is very limited. We discuss related work in phonemic awareness in NLP and other approaches to mitigating the performance divide between languages with different writing systems, and then motivate our own work.

\paragraph{Leveraging Phonemes in Text NLP.} 

Training text-based neural networks on both phonemic and orthographic information has given downstream task performance improvements in mono- and multi-lingual NLP and speech tasks \cite{chen2014joint, bharadwaj-etal-2016-phonologically, liu-etal-2024-translico}. However, such work was mostly limited to smaller LMs or task-specific models (<500M parameters) where parameter and data constraints remain potential obstacles for effective integration \cite{wang2020negative}. On the other hand, as LLMs scale up to many billions of parameters, training-based schemes face limited orthographic-phonemic data and high computational costs, motivating an inference-time approach.

\paragraph{Performance Gaps in Multilinguality.} It was noted in early <500M-parameter LMs like mBERT \cite{devlin2019bert} and XLM-R \cite{conneau2020unsupervised} that performance varied widely across languages \cite{wu-dredze-2020-languages}, an observation that has persisted to modern LMs (see references in \Cref{sec:introduction}). For example, such LMs had greater difficulty in adapting to languages with non-Latin scripts \citep{muller2021being,pfeiffer2021unks}.

Non-phonemic approaches to bridging multilingual gaps via finetuning have involved modified training, synthetic data via translation, and contrastive presentations \cite{zheng2021consistency,kumar2022mucot, yang2022enhancing}. With regards to writing systems, while existing works explored improving transfer capability \cite{fujinuma-etal-2022-match,nguyen2023enhancing, liu-etal-2024-translico}, they also remained limited to small LMs (<2B parameters).
Work on narrowing gaps in prompting and ICL performance have involved cross-lingual chain of thought and demonstrations via translation \citep{qin2023cross,ranaldi-etal-2024-empowering}, but have not to date involved phonology.

\paragraph{Alternate Transcriptions in LLM Prompts.}
Integrating other linguistic knowledge beyond the textual scripts  could lead to more robust and generalizable language models \cite{linzen2020can}.
Even in early multilingual models, using orthographic text as input for adaptation to unseen settings has been shown to provide little gains for \nonlatin~script languages. \citet{muller2021being} proposed using transliteration during finetuning as a scheme to pass from \nonlatin~to \latin~tokens that were at least phonetically similar.
More recently, prior works have proposed utilizing romanization as an augmentation scheme for orthographic text inputs \cite{husain2024romansetu}; their motivation does not mention phonology but rather script and token overlap with common Latin-script languages. However, romanizations might not exist for all languages, limiting the potential adaptation towards truly low-resource languages \cite{doi:10.1080/01434632.2017.1284855}. While prompting is considered, retrieval is not. Finally, Romanization varies widely across languages (e.g., Wade-Giles vs.\ Hanyu Pinyin for Mandarin Chinese), though modern software has improved the selection of crosslingually consistent schemes \cite{hermjakob-etal-2018-box}.

In contrast, in this work, we focus on phonemic IPA transcriptions, which balance a universally applicable transcription system capturing the variability of sounds across languages \cite{mortensen-etal-2016-panphon,bharadwaj-etal-2016-phonologically} while staying largely imputable from written text, unlike full phonetic transcriptions. Transliteration and romanization, which were motivated by gains from passing to \latin~script tokens, may limit benefits to certain groups of languages and only incidentally capture phonology. Furthermore, our use of IPA relies on less-frequent characters that LLMs have seen primarily in phonological contexts; beneficial for in-context reasoning with less spurious connotations. We investigate how phonemic integration in LLM prompting might help improve the downstream task inference performance across \nonlatin~scripts and in comparison with Romanization.

%%%%%%%%%%%%%%%%%%%%%%%%%%%%%%%%%%%%%%%%%%%%%
\begin{figure*}[htbp!]
    \centering
    \includegraphics[width=1\linewidth]{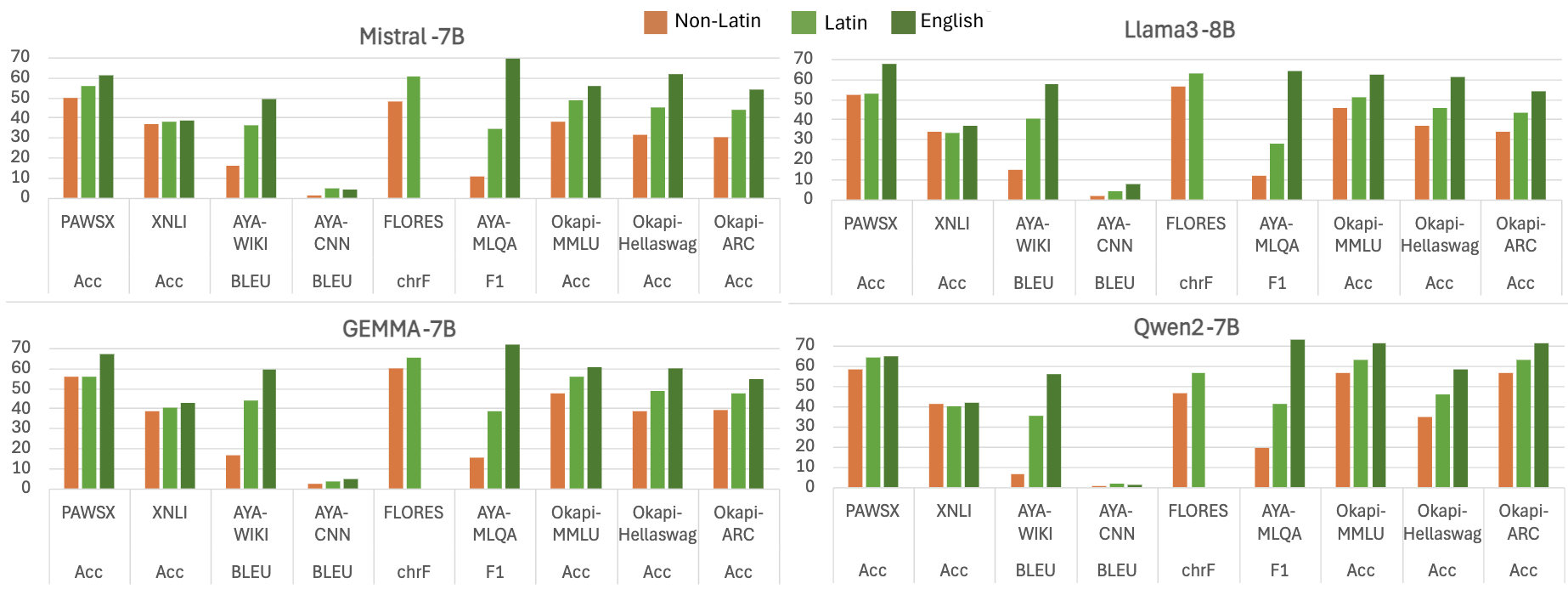}
    \caption{Performance on open-weights LLMs with size around 7B; languages grouped into \nonlatin~scripts, \latin~scripts (excluding English), and English respectively.}
    \label{fig:pilot_study}
\end{figure*}

\section{Is the Written Script Sufficient for Multilinguality?} \label{sec:pilot-study}

While the multilingual discrepancy in performance when comparing \nonlatin~ and \latin~ languages have been studied for smaller Transformer-based LMs, these studies may not fully apply to recent contemporary LLMs with more capacity ($\ge$7B parameters) and training data, thus requiring additional in-depth investigations.

To empirically measure the gap in performance on \latin~versus \nonlatin~scripts as a baseline for our work, we start with a pilot study across 4 \nonlatin~script languages--Hindi (hin), Arabic (arb), Chinese (zho), Japanese (jpn)\footnote{Following \citet{singh2024aya}, we adopt the ISO-639-3 language code abbreviation for conciseness.}---and 6 \latin~script languages---German (deu), French (fra), Dutch (nld), Italian (ita), Portuguese (por), Spanish (spa).
A suite of datasets, grouped by task family (as presented in Table \ref{tab:appendix_pilot_task}), are evaluated, ranging from natural language understanding (NLU), natural language generation (NLG), machine translation (MT), and question answering (QA). The tasks were chosen for having a nearly similar set of aforementioned languages available for evaluation, enabling fair comparisons across tasks and language categories. For this initial suite of evaluations, we evaluated 4 base LLMs in the same weight class: Mistral-7B~\cite{jiang2023mistral}, Llama3-8B~\cite{llama3paper}, Gemma-7B~\cite{team2024gemma} and Qwen2-7B~\cite{yang2024qwen2}. Additional details regarding the coverage of tasks, evaluation metrics, language, LLMs, and experimental setup are provided in Appendix \ref{sec:appendix-pilot-details}.

\input{final_tables/appendix/pilot_dataset}

Our empirical results in \Cref{fig:pilot_study} demonstrate that the performance of \nonlatin~script languages remains inferior to \latin~script languages across LLM families in this weight class. Specifically, our evaluation reveals an average 23\% difference in performance across all tasks, with a marked disparity on generation tasks (reaching approximately 65\%). Notably, while the performance gap is minor on NLU tasks such as PAWS-X and XNLI across the majority of LLMs, the performance of \nonlatin~script languages on NLG tasks such as \wiki~is significantly worse compared to performance on \latin~script languages, especially English. For instance, the performance gaps between \nonlatin~and \latin~script languages are 27, 23, and 5 points on~\wiki,~\mlqa\footnote{The original MLQA did not cover all languages, so for consistency we used \mlqa. We quantify the effects of this in \Cref{sec:appendix-mlqa-discrepancy}.}~and~\flores~with Gemma-7B. With Qwen2-7B, we also observe the significant the gap on~\wiki~(approximately 29 points). We also observe an analogous gap on the Aya-CNN task, though overall performance is worse as the dataset contains more noise and long-context samples quickly went beyond the evaluated LLMs' context windows under few-shot settings. Overall, these findings highlight that the \nonlatin~vs~\latin~language performance gap persists across LLM families and to this day.

To reduce this performance gap, we suggest incorporating phonemic information in model prompting, building on the insights from \citet{ziegler2005reading}. Their ``psycholinguistic grain size theory'' explains that learning to read depends on phonological awareness (the ability to recognize and work with sounds in language), and the complexity of a language's writing system determines how we process it---whether by focusing on letters, syllables, or whole words. Based on this, we believe that adding phonemic information can help LLMs better process \nonlatin~script languages, just as phonological awareness helps language learning for humans. 

%%%%%%%%%%%%%%%%%%%%%%%%%%%%%%%%%%%%%%%%%%%%%
\input{final_tables/main_result/llama3-qwen2-icl-results}

\section{Prompting with Phonemes} \label{sec:phonemic-integration}

These insights focused our experiments on a representative from each task category where different LLM families struggled to achieve similar performance between \latin~and \nonlatin~languages: \wiki~(NLG), \flores~(MT), and question answering (\mlqa) and compare on the same metrics.
Due to the lack of publicly available multilingual corpora with orthographic-phonemic alignments, we adopt the approach of \citet{bharadwaj-etal-2016-phonologically} and \citet{nguyen2023enhancing} to construct our own aligned dataset for evaluation. Specifically, we use Epitran \cite{Mortensen-et-al:2018}, a tool based on linguistic references, to generate IPA transcriptions\footnote{We take Mandarin Chinese pronunciations for ZHO.} for orthographic-only multilingual datasets, setting up the foundation for our phonemic integration explorations with LLMs.

In this work, we explore a series of experiments on phonemic integration with text-based LLMs, improving inference-time performance without the need for pretraining or fine-tuning. In the main text, without the loss of generality, we focus on two LLMs, \llama~\cite{llama3paper} and \qwen~\cite{yang2024qwen2}. Similar results for Mistral and Gemma instruction-tuned variants are in \Cref{sec:appendix-mistral-gemma-variant}. In this section we explore two prompting approaches, \textit{direct} prompting and \textit{ICL retrieval}-based prompting, and find that phonemic integration indeed helps reduce the performance gap between the script categories.
%%%%%%%%%%%%%%%%%%%%%%
\subsection{Phonemic Integration via Direct Prompting} \label{sec:phoneme-direct-prompting}
\input{final_tables/main_result/prompt-variations}
We first study the most straightforward approach to inject phonemic information: direct prompting. More specifically, under the assumption that LLMs might be able to surface internal knowledge about correspondences between script and phonemes effectively, we append the phonemic IPA transcription in the prompt as additional auxiliary information to the original text. We conduct experiments with 0-shot and 3-shot prompting with random samples. Additional details of prompt templates and variations are provided in \Cref{sec:appendix-prompt-template-by-task}.
%%%%%%%%

%%%%%
As shown in \Cref{tab:prompt_variations}, concatenating IPA information to orthographic inputs consistently improves performance across various tasks and evaluation settings, except in the 3-shot setting for~\flores. Given that 0-shot performance on~\flores~is already high with standard prompting, we hypothesize that text-based LLMs can effectively handle \flores~without much benefit from adding IPA information through in-context examples.
%%%%%%%%%%%%%%%%%%%%%%%

\subsection{Phonemic Integration via ICL Retrieval} \label{sec:phoneme-icl-retrieval}

Besides directly injecting IPA information through prompting, we explore the use of phonemic IPA information to enhance retrieval for ICL. Since LLMs are highly sensitive to various prompt formats \cite{lu2024prompts}, we keep the Script+IPA prompt format from \Cref{sec:phoneme-direct-prompting} while leveraging different ICL retrieval strategies in a fixed 3-shot setting.

%%%%%%%%%%%%%%%%%%%%%%%%%%%%%%%%%%%%%%%%%%%%%
%%%%%%%%%
\begin{figure*}[t]
    \centering
   \includegraphics[trim={1.4cm 4.7cm 0.0cm 0.5cm},clip,width=\textwidth]{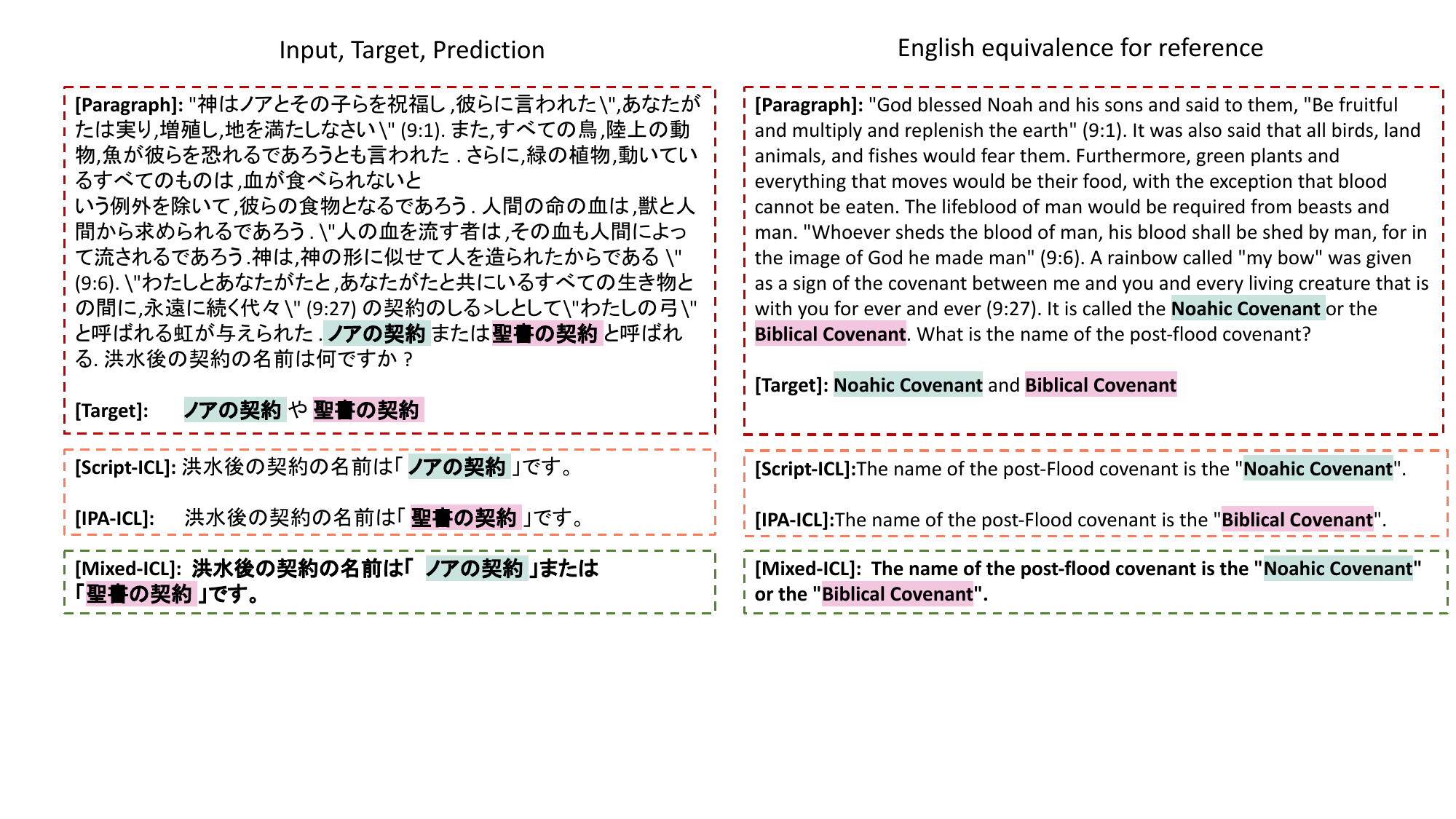}
    \vspace*{-0.5cm}
    \caption{Example for JPN on \mlqa~showing the generated output from LLMs induced by prompts populated by different ICL retrieval methods (\scripticl~vs.\ \ipaicl vs.\ \mixicl). \mixicl~yields a more comprehensive output that is semantically closer to the target than its \ipaicl~and \scripticl~counterparts.}
    \label{fig:case-study-mixed}
\end{figure*} 
Additionally, we compare the impact of few-shot example retrievals based on phonemic and orthographic similarity, against a baseline of randomly retrieved examples (\texttt{Random}). The retrieval methods include orthographic-based matching (\scripticl), IPA-based matching (\ipaicl), and our proposed mixed strategy (\mixicl). In the \mixicl~approach, matching scores are calculated separately with \script~and~\ipa, then averaged for each sample. The top 3 samples are selected after re-ranking by the averaged scores, leveraging both orthographic and phonemic similarity information. Additional in-depth comparisons with other mixing strategies are further explored in Appendix \ref{sec:appendix--mix-strategy}. Due to a lack of phonemic transcription embedders, we focus our studies on using lexical retrieval---namely, BM25 sparse retrieval \citep{trotman2014improvements,luo2023dricl} under each LLM's tokenization---to compute matching scores across for all ICL variants in the main text. However, similar studies on a dense retriever variant (\Cref{sec:appendix-dense-icl}) further validate the effectiveness and flexibility of our method.

We find that the ICL retrieval methods generate strong improvements over the Random baselines for both \llama~and~\qwen~(\Cref{tab:llama3-qwen2-icl-results}). 
For example, in \llama we observe average performance improvements of 1.1 and 0.4 points absolute on \mlqa~with \scripticl~and~\ipaicl, respectively. The gain is further boosted to 2.4 points with our proposed \mixicl~strategy by combining benefits from both \script~and~\ipa. We further analyze our observations and findings with fine-grained analysis on aspects such as the impact of \latin~vs.~\nonlatin~scripts, \scripticl~vs.~\ipaicl~retrieval.

%%%%%%%%%%%%%%%%%%%%%
\subsection{Reducing the Performance Gap between \latin~versus~\nonlatin~Languages}
We observe that the inclusion of IPA information in ICL leads to improved performance for both \latin~and~\nonlatin~script languages, and especially contributes to better performance for \nonlatin~script languages (12.6\%, 1.0\% and 8.6\% relative performance gain over random retrieval for \wiki,~\flores, and \mlqa~ respectively), helping reduce the previously observed performance gap (\Cref{tab:latin-vs-nonlatin-icl-improvement-over-random}). In particular, we find that the \mixicl~strategy contributes to the most gains across all tasks, with the exception of~\flores~on \latin~languages, where the \ipaicl~strategy achieves stronger performance gain. This reinforces our main finding: integrating phonemic information in ICL prompting leads to improved performance for not just \nonlatin~languages, but also \latin~languages, with \nonlatin~languages seeing higher gains. This study also reiterates the broad benefit of IPA as a phonemic representation for improved language support.
%%%%%%%%%%%%%%%%%%%%%%%%%%%%%%%%%%%%%%%%%%%%%
\input{final_tables/ablations/latin-vs-nonlatin-icl-improvement-over-random}
%%%%%%%%%%%%%%%%%%%%%%%
\begin{figure*}[ht]
    \centering
    \includegraphics[trim={0.2cm 5.6cm 0.0cm 0.2cm},clip, width=\textwidth]{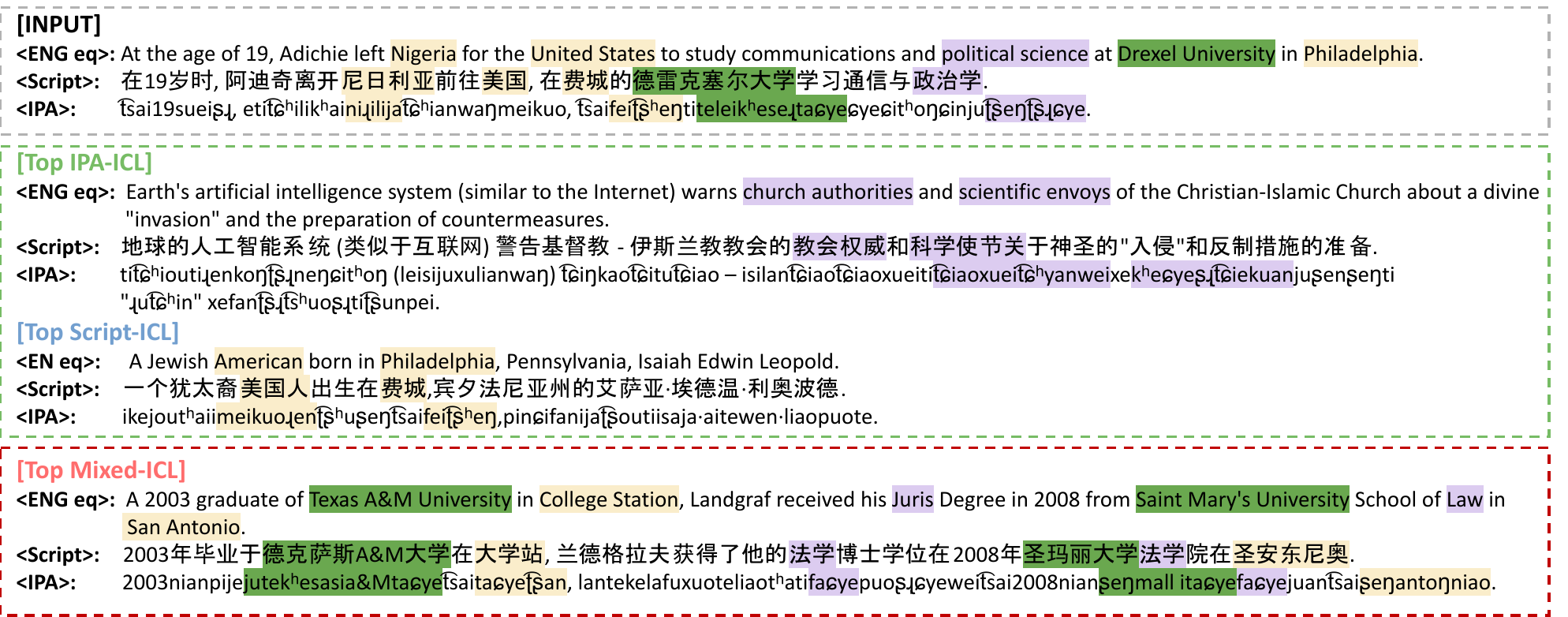}
    \vspace*{-0.5cm}
    \caption{Example of top-1 retrieved samples from different ICL retrieval schemes for the [INPUT] query sample. <ENG eq> denotes the equivalent translation of the target language input in English for readability. <Script> and <IPA> denote the orthographic and corresponding phonemic IPA representation of each sample, whose LLM tokenizations were used for BM25 queries and retrievals. The highlights represent the aligned concepts captured across examples, including \colorbox{figblue}{\textit{university}}, \colorbox{figyellow}{\textit{location}}, and \colorbox{figpurple}{\textit{governance}} respectively (best viewed in color). }
    \label{fig:case-study-top-retrieved}
\end{figure*} 
%%%%%%%%%%%%%%%%%%%%%%%%%%%%%%%%%%%%%%%%%%%%%
\section{How do \script~vs.~\ipa~vs.~\mixed~Work for ICL?}
To gain a better intuitive understanding of different ICL retrievers, we conduct additional qualitative studies by (1) probing the generated output differences when prompting with different ICL retriever methods, and (2) investigating the top retrieved examples from different retrievers. Observing the LLM generations when prompted with different ICL approaches (\Cref{fig:case-study-mixed}), the~\mixicl~strategy shows a more comprehensive answer than its~\scripticl~and~\ipaicl~counterparts, resulting in the answer closest to the ground truth target among the three compared. 

Additionally, when comparing the top retrieved examples from different ICL approaches (\Cref{fig:case-study-top-retrieved}), we observe the top retrieved example from \mixicl~covers the most similar concepts with the original input query; e.g., \textit{Drexel University} vs.\ \textit{Texas A\&M University} (university), \textit{Philadelphia} vs.\ \textit{San Antonio} (location), \textit{political science} vs.\ \textit{law} (governance). This showcases its closer semantic connections with the input query sample as compared to the other two variants, where fewer similar concepts were captured. These observations support our motivations in synthesizing the knowledge from both Script and IPA information via ICL, helping explain the performance improvements observed in \Cref{tab:llama3-qwen2-icl-results}. 

%%%%%%%%%%%%%%%%%%%%%%%%%%%%%%%%%%%%%%%%%%%%

%%%%%%%%%%%%%%%%%%%%%

\section{Analysis} \label{sec:results}

For consistency and conciseness, \textbf{in this section our results are averages over the aforementioned \nonlatin~languages} unless stated otherwise.

%%%%%%%%%

\paragraph{How diverse are ICL examples retrieved with Script vs IPA?} We present \Cref{tab:overlap_icl} to measure the overlap between the top-3 retrieved samples from \scripticl~and \ipaicl. Average overlaps are low, ranging from 7.8\% to 13.5\%, showing that leveraging phonemic information provides distinct ICL retrieval examples from when orthographic information is considered. Keeping performance in mind (\Cref{tab:llama3-qwen2-icl-results}), this observation points to the fact that retrieval with IPA information is a robust and effective tool.
\input{final_tables/ablations/icl-overlap-examples}

%%%%%%%%%

\paragraph{Are both the script and IPA inputs needed for ICL prompting?} Since \ipaicl~and~\scripticl~only consider the corresponding relevant information for selecting ICL examples, the complementary information (Script and IPA respectively) might be deemed unimportant and provide little information for LLM inference. For instance, if using \scripticl, is it possible that only Script information is essential while the IPA information can be left out? Therefore, we conduct additional studies on the impact of removing this possibly unused information. As observed in \Cref{tab:abbrv_remove-unused-info}, including both Script and IPA information yields the best performance consistently across different tasks. This shows the importance of providing comprehensive information regarding both ICL and query samples for LLMs' predictions regardless of the types of information used for ICL retrieval. In addition, we observe a significant performance decrease when Script information is removed from the prompt, aligning with claims from previous work that phonemic information is considered more as an addendum, rather than a replacement, to textual scripts for enhancing downstream task performance for text-based LMs \cite{nguyen2024cori}. While all of the evaluated tasks require script output, the task performance when inputs are limited to phonemic representations remains notably above zero. This suggests that phonemic representations might preserve linguistically informative features and/or LLMs retain implicit script-specific biases regardless of language inputs. Further investigations are needed to rigorously assess the implications in detail. 
\input{final_tables/ablations/abbr_tab/effect-of-removing-unused-info}

\paragraph{Does~\ipaicl~retrieval vary with different tokenizers?} Since text-based LLM tokenizers have seen relatively few IPA inputs, we investigate the potential impact of different tokenizers on \ipaicl. As observed in \Cref{tab:abbrv_tokenizer_impact}, most tokenizers show the benefits of~\ipaicl~over Random baseline. However, we observe that naïve alternate tokenizations for BM25 retrieval---per-character separation (CS) and whitespace separation (WS)---typically perform worse than tokenizers from pre-trained LLMs (14.4\% vs 10.8\% relative performance gain on~\wiki~with \llama~and WS tokenizers respectively). On~\mlqa, WS and CS tokenizers may even hurt task performance, resulting in relative performance decreases of 0.7\% and 3.4\% when compared to the \textit{random} baseline.
\input{final_tables/ablations/abbr_tab/tokenizer_impact}

\paragraph{IPA versus romanization.}
\label{sec:appendix-ipa-vs-roman}

As previously mentioned, romanization is also considered a viable phonemic signal. Despite the appealing of leveraging roman characters to transliterate and/or capture pronunciation for the target languages, romanization is not standardized and heavily language-specific, resulting in various potential romanization schemes for given languages. More importantly, since romanization is only complementary for \nonlatin~script languages, it cannot be utilized to capture phonemic information for \latin~languages, unlike IPA which enhances LLMs multilingual capability for \latin~script languages also.

For completeness, we conduct additional investigations on prompting LLMs with romanization, as similarly done with IPA throughout our work. Following \citet{husain2024romansetu} and \citet{nguyen2024cori}, we used language-specific romanization tools to generate romanization-aligned versions of textual script data. Similar to Section \ref{sec:phonemic-integration}, we explore two different directions (1) Direct prompting with romanization and (2) Enhanced romanization-ICL instead of IPA-ICL.
\input{final_tables/appendix/roman_prompt_variation}

%%%%%%%%%%
\textit{Direct Prompting.} As indicated in \Cref{tab:roman_prompt_variations}, similar to Script+IPA, Script+Roman.\ gains improvements over the Script-only prompting approach in most cases. However, it is unclear whether IPA or romanization serves as a better phonemic signal for this approach. For instance, on~\wiki~task, Script+IPA tends to perform better than Script+Roman, but the opposite might exist on~\flores~task. Hence, we further study ICL based approaches with romanization as well.

%%%%%%%%%%

\textit{ICL retrieval benefits with IPA and romanization.} We evaluate the impact of an IPA-ICL retriever and a romanization-ICL retriever independently from the impact of the prompting variation as follows: (1) Only romanization is used within the prompt as phonemic signal, and (2) Only IPA is used within the prompt. Table \ref{tab:appendix-roman-icl} reveals that our IPA retriever is consistently more effective than the romanization-based counterpart across all of the evaluated tasks regardless of phonemic signals used for prompting, e.g., giving 2.3 and 1.4 BLEU point gaps on \wiki~when prompting with only Romanization and only IPA respectively. With this observation, we focus our main studies on phonemic integrations using IPA as phonemic signals. We leave further in-depth comparative studies between romanization and IPA for future work.
\input{final_tables/appendix/roman_icl_comparison}

\textit{Compound benefits of \ipaicl~and \romanicl.} As both romanization and IPA can be essential complementary information beyond the written scripts, we conduct additional investigations on whether aggregating both information as a comprehensive enhanced ICL approach can further facilitate the multilingual capability of LLMs. Concisely, we evaluate the \allicl~variant, an ICL approach leveraging \textbf{all} information including Script, IPA, and Romanization for ICL retrieval with the aforementioned \mixicl~aggregation mechanism. As observed in \Cref{tab:roman_icl_complement}, our empirical studies demonstrate the complementary benefits of leveraging different additional sources of information for enhanced ICL, leading to the most significant performance improvements across all evaluated downstream tasks.
\input{final_tables/appendix/roman_icl_complement}

\paragraph{Other analyses.} We encourage the reader to see our additional analyses in \Cref{sec:additional-explorations}. Beyond ones already referenced, we also briefly evaluated on large proprietary LLMs (GPT-4 and Mixtral-8x22B-Instruct) in \Cref{sec:appendix-gpt4}, and the effectiveness of finetuning with IPA and using greater numbers of shots in \Cref{{sec:appendix-icl-sft}}.

Overall, our analyses confirm our findings: IPA information is an effective tool, even in different settings, towards better inference-time performance---especially for \nonlatin~languages. 

%%%%%%%%%%%%%%%%%%%%%%%%%%%%%%%%%%%%%%%%%%%%%
\section{Conclusion} \label{sec:conclusion}

We investigated the effect of integrating phonemic information towards improving the multilingual abilities of large-scale language models at inference time. Our pilot study demonstrated that the performance gap between \latin~and~\nonlatin~script languages remains high even on latest state-of-the-art LLMs with an approximate average 23\% difference in performance across all evaluated tasks, and more substantial difference on generation tasks (approx.\ 65\%). 

Motivated by these observations, we proposed introducing phonemic integration in prompting with LLMs. We explored two incorporation mechanisms: direct prompting and ICL retrieval. While we observed performance gains with direct prompting, our empirical study demonstrates that the ICL retrieval provides an even more effective way to improve downstream task performance. The \mixicl~retrieval strategy captures diverse and similar ICL examples versus using textual scripts alone, leading to the best overall performance across multiple tasks, also evidenced in the case studies we present.

We contributed an extensive empirical study on the effect of integrating phonemic information towards improving the performance of contemporary LLMs, reducing the performance gap between \latin~and~\nonlatin~performance. We found that incorporating phonemic information as IPA with few-shot ICL retrieval prompting is an effective method to improve multilingual performance for languages with differing written scripts.

%%%%%%%%%%%%%%%%%%%%%%%%%%%%%%%%%%%%%%%%%%%%%

\section*{Limitations} \label{sec:limitation}

We conducted multiple studies and analyses towards providing a comprehensive report on how phonemic integration with orthographic prompting can improve performance for \nonlatin~script languages, especially towards reducing the performance gap between their \latin~counterparts. However, our study has its limitations.

First, we rely on the external resources for both multilingual evaluation datasets and IPA generation tools to generate the phonemic text input, and thus rely on their quality. To the best of our knowledge, we utilize the best, linguistically informed, IPA generation tool that has been widely adopted by previous works for preprocessing IPA transcriptions \cite{bharadwaj-etal-2016-phonologically, nguyen2023enhancing}. Regarding evaluation datasets, with the goal of future extensions towards more languages, we leverage the most comprehensive multilingual datasets available, including Aya Collections \cite{singh2024aya} and Okapi \cite{lai2023okapi}. We heavily rely on their data quality control protocols for our target languages. However, since the datasets leverage different machine translation tools to generate corpora across a large number of supported languages, the data quality for our targeted \nonlatin~languages might not be optimal. Despite our attempts in data cleaning as mentioned in Appendix \ref{sec:appendix-dataset}, by comparison in cases of available ground-truth translations we did find that the use of machine translation affected absolute model quality, though we view this as orthogonal to the benefits of IPA \cref{sec:appendix-mlqa-discrepancy}

Second, our work is restricted to phonemic integration via prompting. Unlike previous works that explore instruction fine-tuning and continual pre-training concurrently \cite{husain2024romansetu}, we seek to provide in-depth insight into the effect of phonemic integration with text LLMs on downstream task performance as is most immediate to practitioners, i.e., without the additional parameter updates and training objectives of fine-tuning and pretraining paradigms. The study gives foundational guidelines for future fine-tuning approaches for better alignment between phonemic and orthographic signals.  

Third, we explored a simple preliminary \mixicl~strategy to aggregate the benefits from both IPA and Scripts. The promising results not only provide insights into the early exploration of phonemic integration with text-based LLMs but also encourage future works on investigating more effective and dynamic aggregation mechanisms
\cite{ye-ling-2019-multi,nguyen2020dynamic,nguyen2023slot} 
to enhance the benefits further.

Lastly, our study is limited to 4 \nonlatin~and 6 non-English \latin~script languages. However, we ensure that each of the \nonlatin~languages chosen for the study have different written scripts (Table \ref{tab:appendix_pilot_language}), making them diverse and the task of evaluation complex. Extending the study to more languages covering more diverse linguistic groups, writing scripts, language families \cite{nguyen2019cross,singh2024aya} remains a future direction for our research.

%%%%%%%%%%%%%%%%%%%%%%%%%%%%%%%%%%%%%%%%%%%%%

\section*{Ethical Considerations and Societal Impact} \label{sec:ethical-consideration}

Our study is mainly empirical in nature, and does not involve studies with humans. We also utilize publicly available datasets, which we expect to be non-toxic. We plan to make our unique IPA-orthographic aligned data available publicly, and hope that along with our study, it motivates further exploration and research into the potential of including prompting with phonemic information towards better performance.

We believe our study can have a positive impact in NLP by reducing the performance gap of \nonlatin~script languages compared to \latin~script language for LLMs. We hope our work will encourage the exploration of new methods in (and beyond) phonemic integration, towards further reducing the performance gap and improving access for all.

Our work only generates IPA transcription from the publicly available multilingual benchmark datasets. In other words, we do not generate any new content besides the transcriptions of the given textual scripts; hence, our work and its generated data do not pose any potential risks beyond potentially incorrect and opinionated pronunciations.

%%%%%%%%%%%%%%%%%%%%%%%%%%%%%%%%%%%%%%%%%%%%%

\bibliography{references}

%%%%%%%%%%%%%%%%%%%%%%%%%%%%%%%%%%%%%%%%%%%%%
%--- New page for clarity now, will merge later if needed

\appendix

\section{Additional Setup Details} \label{sec:appendix-setup}

%%%%%%%%%%%%%%%%%%%%%
\subsection{Pilot Study Details}
\label{sec:appendix-pilot-details}
In this section, we provide details regarding our ``pilot study'' setup (\Cref{sec:pilot-study}), separated into 3 main sections: (1) Language Coverage, (2) Task Coverage, (3) Experimental Setup.

\input{final_tables/appendix/pilot_language}
\paragraph{Language coverage.} The objective of our pilot evaluation was to compare and contrast performance of contemporary LLMs ($\ge$7B parameters) on \nonlatin~and \latin~languages. English (ENG) performance is considered the upper bound performance. Details of covered languages with their corresponding written scripts are presented in Table \ref{tab:appendix_pilot_language}.

\paragraph{Task coverage.} With the goal of conducting a fair multilingual evaluation of contemporary LLMs, we cover a wide range of tasks where all of the considered \nonlatin- and \latin-script languages are available (Table \ref{tab:appendix_pilot_task}). We separated the tasks into 4 major categories: natural language understanding (NLU), natural language generation (NLG), machine translation (MT) and question answering (QA). Following \citet{singh2024aya,husain2024romansetu}, we leverage standard evaluation metrics for each task type. For NLU, as the PAWS-X \cite{pawsxpaper} and XNLI \cite{xnlipaper} datasets contains both \latin~ and \nonlatin~script languages, we leverage the original ones for our pilot study. In cases where the target languages are not present in the original datasets, we employ internally generated translations from English to the respective languages. For the \flores~dataset where high-quality translations are available across 200 languages \cite{florespaper}, we focused on evaluating the translation capability of LLMs from the target language to English (target $\rightarrow$ ENG). For multiple-choice question answering (QA), we extract the multilingual versions of the MMLU, HellaSwag and ARC datasets accumulated by the Okapi dataset collection \cite{lai2023okapi}.

For NLG and Extractive QA tasks, we leverage the Aya Collection \cite{singh2024aya}, where these tasks are available in 114 languages, including the considered \nonlatin~- and \latin~- languages. Unlike the original multilingual MLQA \cite{mlqapaper} datasets, Aya Collections leverage the English (ENG) splits and generate the corresponding equivalence for other languages via NLLB translation tools \cite{nllb2022}. As translation may degrade performance on certain low-resource languages, including our \nonlatin~languages, direct comparison between previous SOTA baselines and our empirical study might be inappropriate. We quantify the degradation caused by the auto-translated versions of these datasets in Section \ref{sec:appendix-mlqa-discrepancy}.

\paragraph{Experimental setup.}
We conduct our empirical study via the EleutherAI evaluation framework\footnote{\url{https://github.com/EleutherAI/lm-evaluation-harness}} where the default settings and prompts for each task are leveraged. The results are reported in 3-shot settings so that LLMs can follow the output format from the in-context examples and generate appropriate output responses for the given tasks. Random sampling is leveraged to extract 3-shot examples for each given query sample. 
%%%%%%%%%%%%%%%%%%%%%
\subsection{Details of Main Experimental Setup}

%%%%%%%%%%

\paragraph{Prompt templates by task.} \label{sec:appendix-prompt-template-by-task}
In our experiments we observe that slight prompt variability can drastically affect task performance \cite{lu2024prompts}. Additionally, certain challenging tasks such as \mlqa~rely on sufficient task context and prompt templates to execute the given tasks effectively. Therefore, we outline the specific prompts leveraged in our study in \Cref{tab:prompt-templates-by-task}. Consistent prompting eliminates the influence of prompt variations on the observed differences in our empirical studies. Instead, the changes introduced by our individual study can be directly measured through the performance evaluation. The optimal prompt selection and tuning for the best downstream task performance is beyond the scope of our study.

\input{final_tables/appendix/prompt-by-task}

%%%%%%%%%%

\paragraph{Dataset statistics.}
\label{sec:appendix-dataset}
Considering the future extensions of our work towards more languages, we purposely conduct our evaluations on the multilingual datasets covering a wide range of languages. For machine translation, we used~\flores~\cite{florespaper}. For~\wiki~and~\mlqa, we used the versions from the Aya Collection \cite{singh2024aya} covering 114 languages. However, as Aya leveraged NLLB translation tools \cite{nllb2022} to generate translations, the translation data might be noisy due to potential low-quality generations such as <unk> tokens, etc. Therefore, we conduct further quality control steps to filter out the noisy samples. For each task, we take 500 randomly sampled examples to form the testing set ($D_{test}$). We also construct an example pool of 10,000 samples from train\_set ($D_{pool}$) for ICL retrieval where $D_{pool} \cap D_{test} = \varnothing$. 
The sole major exception is 1012 samples for $D_{pool}$ on~\flores~due to the data availability \cite{florespaper}. Additionally, after quality control filtering for the Aya Collection-derived datasets, for MLQA we got $D_{pool}=5192$ samples for HIN. However since $1012 \gg k$ and $5192 \gg k$ where $k$ = 3, we believe the performance of ICL retrieval is not heavily affected. 
BM25-based ICL retrieval extracts the top-k scoring examples from $D_{pool}$ as examples to prompt LLMs, where $k$ is the number of in-context examples for the given query sample. In most of our experiments, we experiment with $k$ = 3 unless stated otherwise.

We acknowledge the Apache 2.0 License from the Aya Collection \cite{singh2024aya} and the MIT license of Epitran \cite{mortensen-etal-2016-panphon} when constructing the orthographic-phonemic aligned datasets and conducting evaluations for the aforementioned tasks. 

%%%%%%%%%%

\paragraph{Implementation and hyperparameters.}
\label{sec:appendix-implementation}

Similar to our pilot study presented in Section \ref{sec:appendix-pilot-details}, we conducted our empirical study via the EleutherAI evaluation harness. The hyperparameters are set similarly to the default configuration for each individual task. We conduct evaluation on \llama~and \qwen~models on two A100 80GB GPUs. Each suite of experiments across all evaluated tasks took approximately 2 hours, resulting in a total of around 10 hours of inference per LLM.

%%%%%%%%%%%%%%%%%%%%%

\section{Additional Experimental Explorations}
\label{sec:additional-explorations}

%%%%%%%%%%%%%%%%%%%%%
\input{final_tables/appendix/llama3-qwen2-dense-icl-results}

\subsection{Experiments with other 7B/8B LLMs}
\label{sec:appendix-mistral-gemma-variant}
Beyond Table \ref{tab:llama3-qwen2-icl-results}, we conduct further evaluation of our enhanced ICL on other mid-sized LLMs, including: Gemma-7B-Instruct \cite{team2024gemma}, Mistral-7B-Instruct \cite{jiang2023mistral}. The empirical results demonstrate the consistent benefits of our proposed enhanced ICL across the mid-sized (7B/8B) LLMs as observed in \Cref{tab:gemma-mistral-icl-results}.  
\input{final_tables/appendix/gemma_mistral_icl}

\input{final_tables/main_result/gpt4-mixtral-icl-results-by-lang}

\subsection{Experiments with GPT-4 and Mixtral}
\label{sec:appendix-gpt4}
We conduct early exploratory study on the proprietary, large-scale production LLMs like GPT-4 \cite{gpt4paper} and \mixtral\footnote{\url{https://mistral.ai/news/mixtral-8x22b/}}, selected for their widespread use in industry, towards understanding whether IPA integration helps in the largest models. As seen in \Cref{tab:gpt4-mixtral-icl-results-by-lang} on 100 samples, even with simple task-agnostic prompting, the performance with these proprietary and production models is not consistent. While further investigation is necessary, it is difficult to ascertain data contamination with these models for all the tasks we study \cite{deng2024investigating,li2024task}, and the performance changes indicate that these models might have data contamination for the tasks we experiment on. We also find some samples moderated by the API, however upon further manual examination, the 6 total samples (out of the 400) focus on factual news topics such as a prison catching on fire, or quotes from politicians on nuclear power. Since there are only a few, we expect the performance impact to be minimal. We reserve further examination and exploration for future work.

%%%%%%%%%%%%%%%%%%%%%

\subsection{Dense ICL retrieval also benefits with IPA} \label{sec:appendix-dense-icl}

Due to space constraints, we focused our studies on BM25 retrieval methods in the main paper, and report dense retrieval results here. For dense ICL retrieval, we selected \textit{paraphrase-xlmr-multilingual-v1} \cite{reimers-2019-sentence-bert} as the Encoder for input query and individual samples in the ICL pools. The max\_context\_length is set to 512. The sentence representation is leveraged for cosine similarity computation between query and all pooling examples to select the final top-k ICL examples. Results are reported in \Cref{tab:llama3-qwen2-dense-icl-results}.

\input{final_tables/appendix/gpt4-mixtral-dense-results}

Consistent with sparse BM25 Retrieval results reported in \Cref{sec:phoneme-icl-retrieval}, we observe that our \texttt{Mixed} strategy outperforms \texttt{Script} and \texttt{IPA} based ones on all downstream tasks. Similar to observed performance with open-source LLMs, GPT-4 and Mixtral 8x22B also show a trend where the \texttt{Mixed} strategy outperforms most others. These observations imply that ICL retrieval benefits from looking at both orthographic and phonemic information for better example selection, guiding the LLMs towards desired generation output for downstream tasks.

%%%%%%%%%

\input{final_tables/ablations/abbr_tab/detailed_mixing_variations}
\subsection{Impact of Different Mixing Strategies on ICL}
\label{sec:appendix--mix-strategy}
Besides our \texttt{Mixed} strategy, there exist different approaches towards aggregating information from Script and IPA ICL retrieval. We specifically consider 4 different approaches: (1) using the \texttt{Harmonic Mean} to calculate the mixed score for each pool example (2) \texttt{Split-Half}: We retrieve top-(k//2) examples from Script and IPA separately, then aggregate them together to form the final top-k ICL samples. Within this approach, we evaluate 3 different potential ordering to aggregate ICL examples from different sources, including (a) Script+IPA, (b) IPA+Script, (c) Random Shuffle. This approach requires even k-shot samples, (3) \texttt{Divide-Conquer}: After sorting and retrieving top-k samples for both IPA and Script, we concatenate the corresponding scores to form top-2k samples. These samples are then ranked and filtered down to the top-k samples again based on their corresponding BM25 scores, (4) \texttt{Append}: We simply concatenate the scores from the two approaches and retrieve the top-k highest score as the final selected examples. Unlike previous approaches, this approach can possibly result in similar ICL examples being selected twice in the top-k samples. 

Based on our empirical study in \Cref{tab:abbrv_abl_mixing_variations}, our \texttt{Mixed} strategy yields the best performance across our evaluated tasks. \texttt{Split-Half} aggregation is limited to even-shot ICL samples and can potentially suffer from ordering sensitivity, leading to variable evaluation performance \cite{lu-etal-2022-fantastically}.  

%%%%%%%%%%%%%%%%%%%%%

\input{final_tables/ablations/abbr_tab/icl_sft_table}
\subsection{Possible Future Studies with SFT and ICL}
\label{sec:appendix-icl-sft}
For better understanding of the enhanced ICL as compared to continual pre-training or Supervised Fine-tuning (SFT), we conduct additional studies in which we fine-tune \llama~model with additional multilingual M2Lingual dataset \cite{maheshwary2024m2lingual}.  
We extracted and FT the \llama~model on our targeted \nonlatin~languages. Our SFT training takes around 4 hours on five A100 (80GB) GPUs. 
As indicated in \Cref{tab:abbrv_icl_sft}, without additional multilingual training data for the targeted languages, our BM25-Mixed ICL outperforms attempts in FT LLMs with additional multilingual dataset when the number of ICL examples reach 10 shots. This study reveals two essential implications: (1) High-quality ICL selection not only saves the computational cost of additional training but also quickly integrates rare knowledge of IPA with LLMs, (2) Naively SFT LLMs with phonemic-orthographic data might not be sufficient to extract the alignment between IPAs and scripts, emphasizing the goal of our work in gaining deeper understanding of IPAs integration with LLMs via prompting before FT is executed.

%%%%%%%%%%%%%%%%%%%%%
\section{Detailed Results on \latin~Languages}\label{sec:appendix-latin-icl-performance}

For further clarity of \Cref{sec:results}, we provide additional details of the performance across the evaluated tasks on \latin~ languages, including: German (deu), French (fra), Spanish (spa) and Portuguese (por) in \Cref{tab:llama3-latin-results}. These results are leveraged to calculate the relative performance gain as demonstrated in \Cref{tab:latin-vs-nonlatin-icl-improvement-over-random}.

\section{Aya-MLQA Performance Analysis}
\label{sec:appendix-mlqa-discrepancy}
As observed in Table \ref{tab:llama3-qwen2-icl-results}, our empirical studies yield less competitive performance than the originally reported performance of fine-tuned Pre-trained Language Models (PLMs) on MLQA dataset \cite{mlqapaper}. We hypothesize the discrepancy is mostly caused by the issue of dataset quality differences between the original MLQA (\orimlqa) and MLQA from Aya Collection (\ayamlqa). 

\paragraph{Data quality difference.} Our major objective in leveraging Aya dataset \cite{singh2024aya} is the broad language coverage of up to 102 languages, allowing for the further investigation beyond our 4 targeted languages. However, multilingual versions of Aya datasets across tasks are generated via off-the-shelf NLLB machine translation \cite{nllb2022}, which can be prone to errors. Therefore, the dataset quality between \ayamlqa~and \orimlqa~might be different, leading to incomparable performance between ours and previously reported PLMs performance. We further validate our hypothesis via empirical evaluation of our enhanced ICL on \orimlqa~and \ayamlqa~as demonstrated in Table \ref{tab:appendix_aya_ori_mlqa_compare}. More specifically, we observe approximately 22.71 absolute F1 points between \orimlqa~and \ayamlqa. Additionally, our observed performance is on par with the originally reported mBERT FT.    
\input{final_tables/appendix/promptopt_quality_mlqa_compare}
\input{final_tables/appendix/llama3-latin-performance}

\end{document}

%% file: final_tables/appendix/pilot_dataset.tex
\addtolength{\tabcolsep}{-4pt}
\begin{table}[htbp!]
\centering
\resizebox{\columnwidth}{!}{%
\begin{tabular}{@{}ccc@{}}
\toprule 
 Task & Metrics & Dataset Name \\ \midrule

& & PAWS-X \cite{pawsxpaper} \\
\multirow{-2}{*}{ \begin{tabular}[c]{@{}c@{}} NLU \end{tabular}} & \multirow{-2}{*}{ \begin{tabular}[c]{@{}c@{}} Accuracy\end{tabular}}  & XNLI \cite{xnlipaper} \\
\midrule
& & \wiki~\cite{wikipaper, singh2024aya} \\
\multirow{-2}{*}{ \begin{tabular}[c]{@{}c@{}} NLG \end{tabular}} & \multirow{-2}{*}{ \begin{tabular}[c]{@{}c@{}} BLEU \\ \cite{papineni2002bleu}\end{tabular}}  & Aya-CNN \cite{cnndailypaper,singh2024aya} \\
\midrule
& & \\
\multirow{-2}{*}{ \begin{tabular}[c]{@{}c@{}} MT \end{tabular}} & \multirow{-2}{*}{ \begin{tabular}[c]{@{}c@{}} chrF \\ \cite{popovic2015chrf} \end{tabular}}  & \multirow{-2}{*}{ \begin{tabular}[c]{@{}c@{}} \flores~\cite{florespaper} \end{tabular}}\\
\midrule
& F1  &  \mlqa~\cite{mlqapaper,singh2024aya}\\
& & Okapi-MMLU \cite{mmlupaper,lai2023okapi}  \\
\multirow{-2}{*}{QA} & \multirow{-2}{*}{\begin{tabular}[c]{@{}c@{}} \\ Accuracy \end{tabular}}& Okapi-HellaSwag \cite{hellaswagpaper,lai2023okapi} \\
& & Okapi-ARC \cite{arcpaper,lai2023okapi} \\ 

\bottomrule
\end{tabular}%
}
\caption{Task coverage and evaluation metrics for our baselines, including natural language understanding (NLU), natural language generation (NLG), machine translation (MT) and question answering (QA) tasks.}
\label{tab:appendix_pilot_task}
\end{table}
\addtolength{\tabcolsep}{6pt}

%% file: final_tables/main_result/llama3-qwen2-icl-results.tex
\begin{table*}[htbp!]
\centering
\resizebox{\linewidth}{!}{%
\begin{tabular}{@{}cccccccccccccccc}
\toprule
\multirow{3}{*}{\begin{tabular}[c]{@{}c@{}}Llama3-\\  8B-Instruct\end{tabular}} & \multicolumn{5}{c}{\wiki~- BLEU ($\uparrow$)} & \multicolumn{5}{c}{\flores~- chrF ($\uparrow$)} & \multicolumn{5}{c}{\mlqa~- F1 ($\uparrow$)} \\ \cmidrule(lr){2-6} \cmidrule(lr){7-11} \cmidrule(lr){12-16}  
 & \multirow{2}{*}{0-shot} & \multirow{2}{*}{Random} & \multicolumn{3}{c}{BM25} & \multirow{2}{*}{0-shot} & \multirow{2}{*}{Random} & \multicolumn{3}{c}{BM25} & \multirow{2}{*}{0-shot} & \multirow{2}{*}{Random} & \multicolumn{3}{c}{BM25} \\ \cmidrule(lr){4-6} \cmidrule(lr){9-11} \cmidrule(l){14-16} 
 &  &  & Script & IPA & Mixed &  &  & Script & IPA & Mixed &  &  & Script & IPA & Mixed \\ \midrule
HIN & 3.90 & 37.93 & 39.55 & 40.56 & 40.42 & 52.90 & 58.46 & 58.78 & 58.98 & 58.75 & 35.67 & 47.48 & 47.85 & 47.82 & 49.30 \\
ARB & 4.78 & 26.02 & 28.21 & 28.01 & 27.96 & 48.15 & 59.26 & 59.00 & 59.88 & 60.04 & 19.41 & 32.34 & 31.93 & 30.30 & 30.49 \\
ZHO & 1.50 & 3.79 & 6.52 & 6.51 & 6.84 & 48.71 & 56.10 & 56.01 & 56.00 & 56.32 & 11.17 & 12.11 & 13.38 & 13.70 & 14.38 \\
JPN & 1.26 & 1.26 & 4.27 & 3.89 & 4.18 & 48.90 & 53.67 & 54.22 & 53.89 & 54.66 &17.10 & 21.33 & 24.34 & 23.09 & 28.82 \\ \midrule
Average & 2.86 & 17.25 & 19.64 & 19.74 & \cellcolor{darkspringgreen!20} \textbf{19.85} & 49.67 & 56.87 & 57.00 & 57.19 & \cellcolor{darkspringgreen!20} \textbf{57.44} & 20.84 & 28.32 & 29.38 & 28.73 & \cellcolor{darkspringgreen!20} \textbf{30.75} \\ \midrule
\multicolumn{1}{l}{} &  &  &  &  &  &  &  &  &  &  &  & \multicolumn{1}{l}{} &  &  &  \\ \midrule
\multirow{3}{*}{\begin{tabular}[c]{@{}c@{}}Qwen2-\\  7B-Instruct\end{tabular}} & \multicolumn{5}{c}{\wiki~- BLEU ($\uparrow$)} & \multicolumn{5}{c}{\flores~- chrF ($\uparrow$)} & \multicolumn{5}{c}{\mlqa~- F1 ($\uparrow$)} \\ \cmidrule(lr){2-6} \cmidrule(lr){7-11} \cmidrule(lr){12-16}  
 & \multirow{2}{*}{0-shot} & \multirow{2}{*}{Random} & \multicolumn{3}{c}{BM25} & \multirow{2}{*}{0-shot} & \multirow{2}{*}{Random} & \multicolumn{3}{c}{BM25} & \multirow{2}{*}{0-shot} & \multirow{2}{*}{Random} & \multicolumn{3}{c}{BM25} \\ \cmidrule(lr){4-6} \cmidrule(lr){9-11} \cmidrule(l){14-16} 
 &  &  & Script & IPA & Mixed &  &  & Script & IPA & Mixed &  &  & Script & IPA & Mixed \\ \midrule
HIN & 20.74 & 32.05 & 34.67 & 34.07 & 34.79 & 57.56 & 56.99 & 57.39 & 57.05 & 57.60 & 28.04 & 46.46 & 47.16 & 46.48 & 46.79 \\
ARB & 9.59 & 11.34 & 11.64 & 11.60 & 13.58 & 61.04 & 61.45 & 61.50 & 61.77 & 62.07 & 18.09 & 33.11 & 34.43 & 33.42 & 34.75 \\
ZHO & 2.26 & 1.41 & 1.83 & 2.24 & 1.91 & 58.31 & 58.53 & 58.01 & 58.42 & 58.91 & 7.49 & 8.30 & 9.91 & 8.86 & 10.03 \\
JPN & 2.05 & 2.13 & 2.53 & 2.43 & 2.84 & 55.80 & 55.94 & 56.07 & 55.87 & 56.93 & 14.12 & 22.35 & 22.33 & 22.28 & 22.57  \\ \midrule
Average & 8.66 & 11.73 & 12.67 & 12.59 & \cellcolor{darkspringgreen!20} \textbf{13.28} & 58.18 & 58.23 & 58.24 & 58.28 & \cellcolor{darkspringgreen!20} \textbf{58.88} & 16.93 & 27.55 & 28.46 & 27.76 & \cellcolor{darkspringgreen!20} \textbf{28.54} \\ \bottomrule
\end{tabular}%
}
\caption{Llama3-8B-Instruct and Qwen2-7B-Instruct 3-shot results on non-Latin script languages using BM25 retrieval with \texttt{Random}, \scripticl, \ipaicl, and \mixicl~strategies. 0-shot results included for reference. After averaging across languages, our proposed mixed retrieval strategy outperforms all other methods on all tasks.}
\label{tab:llama3-qwen2-icl-results}
\end{table*}

%% file: final_tables/main_result/prompt-variations.tex
\begin{table}[htbp!]
\centering
\resizebox{\columnwidth}{!}{%
\begin{tabular}{@{}clcccc@{}}
\toprule
 &  & \multicolumn{2}{c}{{\llama}} & \multicolumn{2}{c}{\qwen} \\ \cmidrule(lr){3-4}  \cmidrule(lr){5-6} 
\multirow{-2}{*}{Task} & \multirow{-2}{*}{ \begin{tabular}[c]{@{}c@{}} Prompt \\ Variation \end{tabular}} & 0-shot & 3-shot & 0-shot & 3-shot \\ \midrule
 & Script & 2.45 & \textbf{17.76} & 4.73 & 8.3 \\
\multirow{-2}{*}{ \begin{tabular}[c]{@{}c@{}} \wiki \\ (BLEU)\end{tabular}} & Script + IPA & \textbf{2.86} & 17.25 & \textbf{8.66} & \textbf{11.73} \\ \midrule
 & Script & 38.17 & \textbf{57.06} & 57.96 & \textbf{58.45} \\
\multirow{-2}{*}{ \begin{tabular}[c]{@{}c@{}} \flores \\ (chrF) \end{tabular}} & Script + IPA & \textbf{49.67} & 56.87 & \textbf{58.18} & 58.23 \\ \midrule
 & Script & 17.31 & 20.84 & 15.37 & 27.32 \\
\multirow{-2}{*}{ \begin{tabular}[c]{@{}c@{}} \mlqa \\ (F1) \end{tabular}} & Script + IPA & \textbf{20.84} & \textbf{28.32} & \textbf{16.93} & \textbf{27.55} \\ \bottomrule
\end{tabular}%
}
\caption{Effect of prompting with orthographic and/or phonemic information on Llama3-8B-Instruct and Qwen2-7B-Instruct models. 
}
\label{tab:prompt_variations}
\end{table}

%% file: final_tables/ablations/latin-vs-nonlatin-icl-improvement-over-random.tex
\begin{table}[htbp!]
\centering
\small
% \resizebox{\columnwidth}{!}{%
\begin{tabular}{lccc@{}}
\toprule
\multicolumn{1}{c}{\wiki} & \scripticl & \ipaicl & \mixicl \\ \midrule
\latin & +12.07\% & +12.08\% & \textbf{+12.57\%} \\
\nonlatin & +13.84\% & +14.45\% & \textbf{+15.07\%} \\ \midrule\midrule
\multicolumn{1}{c}{\flores}  & \scripticl & \ipaicl & \mixicl \\ \midrule
\latin & +0.33\% & \textbf{+0.51\%} & +0.37\% \\
\nonlatin & +0.23\% & +0.55\% & \textbf{+1.00\%} \\ \midrule\midrule
\multicolumn{1}{c}{\mlqa}  & \scripticl & \ipaicl & \mixicl \\ \midrule
\latin & +0.60\% & +4.09\% & \textbf{+4.58\%} \\
\nonlatin & +3.74\% & +1.45\% & \textbf{+8.58\%} \\
\bottomrule
\end{tabular}%
% }
\caption{Relative performance improvements using different ICL retrieval methods when compared to \texttt{Random} with Llama3-8B-Instruct. See \Cref{sec:appendix-latin-icl-performance} for the results on \latin~languages used for these computations.}
\label{tab:latin-vs-nonlatin-icl-improvement-over-random}
\end{table}

%% file: final_tables/ablations/icl-overlap-examples.tex
\begin{table}[htbp!]
\centering
\resizebox{\columnwidth}{!}{%
\begin{tabular}{@{}cccccc@{}}
\toprule

Task & HIN & ARB & ZHO & JPN & \textbf{Avg} \\ \midrule
\wiki & 11.13\% & 8.33\% & 6.73\% & 5.07\% & \textbf{7.82\%} \\
\flores & 11.07\% & 8.73\% & 6.80\% & 5.60\% & \textbf{8.05\%} \\
\mlqa & 15.00\% & 8.93\% & 23.40\% & 6.47\% & \textbf{13.45\%} \\
\bottomrule
\end{tabular}%
}
\caption{Percentage of overlapping retrieved ICL examples from \scripticl~and \ipaicl~ Retrieval methods (\textit{\scripticl~$\cap$~\ipaicl}) for evaluated tasks on \llama~model.}
\label{tab:overlap_icl}
\end{table}

%% file: final_tables/ablations/abbr_tab/effect-of-removing-unused-info.tex
\begin{table}[htbp!]
\centering
\resizebox{\columnwidth}{!}{%
\begin{tabular}{@{}cccccc@{}}
\toprule
\multirow{2}{*}{Task} & \multirow{2}{*}{Metric} & \multicolumn{2}{c}{\scripticl} & \multicolumn{2}{c}{\ipaicl} \\ \cmidrule(lr){3-4} \cmidrule(lr){5-6}
 &  & w/o IPA & w/ IPA & w/o Script & w/ Script \\ \midrule
\wiki & BLEU & 17.42 & \textbf{19.64} & 12.90 & \textbf{19.74} \\
\flores & chrF & \textbf{58.12} & 57.00 & 21.13 & \textbf{57.19} \\
\mlqa & F1 & 28.59 & \textbf{29.38} & 3.70 & \textbf{28.73} \\
\bottomrule
\end{tabular}%
}
\caption{Effect of removing unused information from ICL when prompting Llama3-8B-Instruct; ``w/o'' means leaving that field blank, keeping prompts identical otherwise (\Cref{sec:appendix-prompt-template-by-task}) the same (\texttt{<script input>: ``XYZ''{\textbackslash}n<ipa input>: {\textbackslash}n''}).}
\label{tab:abbrv_remove-unused-info}
\end{table}

%% file: final_tables/ablations/abbr_tab/tokenizer_impact.tex
\begin{table}[htbp]
\centering
\resizebox{\columnwidth}{!}{%
\begin{tabular}{cccc}
\toprule
Tokenizer Type & \wiki~(BLEU) & \flores~(chrF) & \mlqa~(F1) \\ \midrule
n/a (Baseline) & 17.25 & 56.87 & 28.32 \\ \cmidrule(lr){1-4}
WS & 19.47 \textcolor{darkspringgreen}{(↑12.83\%)} & 56.58 \textcolor{brickred}{($\downarrow$ 0.51\%)} & 28.11 (\textcolor{brickred}{$\downarrow$0.74\%)} \\
CS & 19.11 \textcolor{darkspringgreen}{(↑10.75\%)} & 56.78 \textcolor{brickred}{($\downarrow$ 0.16\%)} & 27.37 \textcolor{brickred}{($\downarrow$3.35\%)} \\ \cmidrule(lr){1-4}
Dense & 19.19 \textcolor{darkspringgreen}{(↑11.22\%)} & 56.95 \textcolor{darkspringgreen}{($\uparrow$ 0.13\%)} & 28.62 \textcolor{darkspringgreen}{(↑1.06\%)} \\ 
Qwen2-Inst & 19.64 \textcolor{darkspringgreen}{(↑13.82\%)} & 57.12 \textcolor{darkspringgreen}{($\uparrow$ 0.44\%)} & \textbf{29.32 \textcolor{darkspringgreen}{(↑3.53\%)}} \\
\textbf{Llama3-Inst} & \textbf{19.74 \textcolor{darkspringgreen}{(↑14.44\%)}} & \textbf{57.19 \textcolor{darkspringgreen}{(↑0.55\%)}} & 28.73 \textcolor{darkspringgreen}{(↑1.45\%)} \\
\bottomrule
\end{tabular}
}
\caption{Impact of different tokenizations on \ipaicl~performance. WS denotes whitespace separation originally proposed by BM25 algorithms \cite{trotman2014improvements} in the context of English. CS denotes the tokenization in which each character (excluding white space) is treated as a single token. \textit{n/a} denotes the \texttt{Random} sampling baseline, for which tokenization is not applicable.}
\label{tab:abbrv_tokenizer_impact}
\end{table}

%% file: final_tables/appendix/roman_prompt_variation.tex
\begin{table}[htbp!]
\centering
\resizebox{\columnwidth}{!}{%
\begin{tabular}{@{}clcccc@{}}
\toprule
 &  & \multicolumn{2}{c}{{Llama3-8B-Instruct}} & \multicolumn{2}{c}{Qwen2-7B-Instruct} \\ \cmidrule(lr){3-4}  \cmidrule(lr){5-6} 
\multirow{-2}{*}{Task} & \multirow{-2}{*}{ \begin{tabular}[c]{@{}c@{}} Variation \end{tabular}} & 0-shot & 3-shot & 0-shot & 3-shot \\ \midrule
 & Script & 2.45 & \textbf{17.76} & 4.73 & 8.3 \\
\multirow{-2}{*}{ \begin{tabular}[c]{@{}c@{}} \wiki \\ (BLEU)\end{tabular}} & Script + IPA & \textbf{2.86} & 17.25 & \textbf{8.66} & 11.73 \\ 
& Script + Roman & 1.64 & 17.41 & 5.82 & \textbf{13.30} \\
\midrule
 & Script & 38.17 & 57.06 & 57.96 & \textbf{58.45} \\
\multirow{-2}{*}{ \begin{tabular}[c]{@{}c@{}} \flores \\ (chrF) \end{tabular}} & Script + IPA & 49.67 & 56.87 & \textbf{58.18} & 58.23 \\ 
& Script + Roman & \textbf{51.26} & \textbf{57.76} & 57.23 & 58.41 \\
\midrule
 & Script & 17.31 & 20.84 & 15.37 & 27.32 \\
 
\multirow{-2}{*}{ \begin{tabular}[c]{@{}c@{}} \mlqa \\ (F1) \end{tabular}} & Script + IPA & 20.84 & 28.32 & \textbf{16.93} & \textbf{27.55} \\ 
& Script + Roman & \textbf{20.95} & \textbf{29.79} & 13.20 & 27.00 \\
\bottomrule
\end{tabular}%
}
\caption{Effect of prompting with both IPA and Romanization as phonemic information for Llama3-8B-Instruct and Qwen2-7B-Instruct models.}
\label{tab:roman_prompt_variations}
\end{table}

%% file: final_tables/appendix/roman_icl_comparison.tex
\begin{table}[htbp!]
\resizebox{\columnwidth}{!}{%
\begin{tabular}{cccccc}
\toprule
\multirow{2}{*}{Task} & \multirow{2}{*}{Metric} & \multicolumn{2}{c}{Roman.\ Prompt Only} & \multicolumn{2}{c}{IPA Prompt Only} \\
 \cmidrule(lr){3-4}  \cmidrule(lr){5-6}  
 & & \romanicl & \ipaicl & \romanicl & \ipaicl \\ \midrule 
\wiki & BLEU & 16.95 & \textbf{19.28} & 18.35 & \textbf{19.74} \\
\flores & chrF & 57.79 & \textbf{58.21} & 56.92 & \textbf{57.44} \\
\mlqa & F1 & 30.49 & \textbf{30.69} &  28.46 & \textbf{28.73} \\ \bottomrule
\end{tabular}
}
\caption{Effects of \romanicl~and \ipaicl~retrieval under both romanization-only and IPA-only prompting. BM25 and \llama~are used for all entries.
\label{tab:appendix-roman-icl}
}
\end{table}

%% file: final_tables/appendix/roman_icl_complement.tex
\begin{table}[htbp!]
\centering
\resizebox{\columnwidth}{!}{%
\begin{tabular}{@{}cc|cccc@{}}
\toprule

Task & Random & \scripticl & \ipaicl & \romanicl & \textbf{\allicl} \\ \midrule
\wiki & 16.90 & 19.15 & 19.40 & 19.17 & \textbf{19.67} \\
\flores & 56.23 & 56.68 & 56.85 & 56.56 & \textbf{57.03} \\
\mlqa & 28.32 & 29.38 & 28.73 & 28.46 & \textbf{30.84} \\

\bottomrule
\end{tabular}%
}
\caption{Benefits of different enhanced ICL approaches for the evaluated tasks on \llama~model versus \texttt{Random} sampling. \allicl~denotes the combination of \scripticl,~\ipaicl~ and~\romanicl~via the same aggregation scheme as \mixicl.}
\label{tab:roman_icl_complement}
\end{table}

%% file: final_tables/appendix/pilot_language.tex
\begin{table}[htbp!]
\centering
\resizebox{\columnwidth}{!}{%
\begin{tabular}{@{}cccc@{}}
\toprule
Category & Language Name & ISO Language & Writing Script \\ \midrule
\multirow{4}{*}{ \begin{tabular}[c]{@{}c@{}} \nonlatin \end{tabular}} & Arabic & ARB & Arabic  \\
& Hindi  & HIN & Devanagari \\ 
& Simplified Chinese & ZHO & Hanzi \\
& Japanese & JPN & Katakana, Hiragana, Kanji \\
\midrule
\multirow{6}{*}{ \begin{tabular}[c]{@{}c@{}} \latin \end{tabular}} & German & DEU & Latin \\
& French & FRA & Latin \\
& Italian & ITA & Latin \\
& Dutch & NLD & Latin \\
& Portuguese & POR & Latin \\
& Spanish & SPA & Latin \\

\midrule
English & English & ENG & Latin \\

\bottomrule
\end{tabular}%
}
\caption{Details of language coverage in the pilot study evaluations. English is a Latin-script language, though we keep it separate due to its primary and disproportionate presence in LM training.}
\label{tab:appendix_pilot_language}
\end{table}

%% file: final_tables/appendix/prompt-by-task.tex
\begin{table}[htbp!]
\centering
\scriptsize
\begin{tabular}{p{1.35cm}p{5.65cm}@{}}
\toprule
 Task & Prompt Template \\ \midrule
 
\wiki & \begin{tabular}{@{}p{5.8cm}@{}}\begin{flushleft}\texttt{You are an expert in Text Simplification task. Given the <script input>, generate a more complex version of the given input. You are also given the IPA phonemic transcription as supplementary information. The format will be as follows:} \\
\texttt{\#\# Input: \{\{input\}\}} \\
\texttt{\#\# Input (in IPA): \{\{input (in IPA)\}\}}  \\
\texttt{\#\# Answer: \{\{answer\}\}} \\
\par\medskip
\texttt{The Answer must be a complete sentence with correct grammatical structure of the target language. The Answer must be in the target language script. Follow the examples if given.}\end{flushleft}
\end{tabular} \\ \midrule

\begin{tabular}[l]{@{}p{1.35cm}@{}}  \flores \end{tabular} & \begin{tabular}{@{}p{5.8cm}@{}}\begin{flushleft}\texttt{Given the <script input> and IPA <ipa input>, translate the <script input> to English. Return one single answer. Do not provide explanations. The format is as follows:} \\ 
\texttt{<script input>: \{\{input\}\}} \\
\texttt{<ipa input>: \{\{input\_ipa\}\}} \\ 
\texttt{<script output>: \{\{script\_output\}\}} \\
\par\medskip
\texttt{The Answer must be in the target language script. Follow the examples if given.}\end{flushleft}
\end{tabular} \\ \midrule

\mlqa & \begin{tabular}{@{}p{5.8cm}@{}}
\begin{flushleft}\texttt{You are an expert in extractive question answering. You will be given Context and a Question (Context + Question) and you must generate at most one Answer, based only on the information in the Context + Question. You are also given the IPA phonemic transcriptions of the equivalent Context + Question (in IPA) as supplementary information. The overall format will be:} \\ 
\texttt{\#\# Context + Question: \{\{context+question\}\}} \\

\texttt{\#\# Context + Question (in IPA): \{\{context+question (in IPA)\}\}}\\
\texttt{\#\# Answer: \{\{answer\}\}} \\
\par\medskip
\texttt{The Answer must appear verbatim in the Context + Question. The answer must be in the target language script. If the Question cannot be answered based on the Context, you will output "unanswerable". Follow the examples if given.}\end{flushleft}
\end{tabular} \\ \bottomrule
\end{tabular}%
\caption{Prompt templates by task. \{\{$\cdot$\}\} denotes the corresponding information for individual samples.}
\label{tab:prompt-templates-by-task}
\end{table}

%% file: final_tables/appendix/llama3-qwen2-dense-icl-results.tex
\begin{table}[htbp!]
\centering
\resizebox{\columnwidth}{!}{%
\begin{tabular}{@{}lccccccccc@{}}
\toprule
 & \multicolumn{3}{c}{\wiki} & \multicolumn{3}{c}{\flores} & \multicolumn{3}{c}{\mlqa} \\ \cmidrule(l){2-10} 
\multicolumn{1}{c}{\multirow{-2}{*}{\begin{tabular}[c]{@{}c@{}}Llama-3-8B \\ Instruct\end{tabular}}} & Script & IPA & Mixed & Script & IPA & Mixed & Script & IPA & Mixed \\ \midrule
HIN & 39.86 & 39.88 & 39.44 & 58.59 & 58.63 & 59.16 & 48.47 & 49.41 & 49.25 \\
ARB & 26.51 & 26.90 & 26.83 & 59.79 & 59.40 & 59.78 & 32.49 & 33.46 & 33.29 \\
ZHO & 5.09 & 4.35 & 5.04 & 56.40 & 55.87 & 56.46 & 12.81 & 12.61 & 13.00 \\
JPN & 2.71 & 2.64 & 2.92 & 53.65 & 54.30 & 53.69 & 27.88 & 26.59 & 28.28 \\ \midrule
Average & 18.54 & 18.44 & \textbf{18.56} & 57.11 & 57.05 & \textbf{57.27} & 30.41 & 30.52 & \textbf{30.96}\\ \midrule
 &  &  &  &  &  &  &  &  &  \\ \midrule
 & \multicolumn{3}{c}{\wiki} & \multicolumn{3}{c}{\flores} & \multicolumn{3}{c}{\mlqa} \\ \cmidrule(l){2-10} 
\multicolumn{1}{c}{\multirow{-2}{*}{\begin{tabular}[c]{@{}c@{}}Qwen2-7B \\ Instruct\end{tabular}}} & Script & IPA & Mixed & Script & IPA & Mixed & Script & IPA & Mixed \\ \midrule
HIN & 32.43 & 33.64 & 34.28 & 57.42 & 57.29 & 57.81 & 46.56 & 46.98 & 47.59 \\
ARB & 11.47 & 11.49 & 13.04 & 61.94 & 62.01 & 61.79 & 33.30 & 33.24 & 33.92 \\
ZHO & 1.23 & 1.43 & 1.88 & 58.52 & 58.42 & 58.66 & 8.90 & 8.06 & 9.65 \\
JPN & 1.91 & 2.15 & 2.60 & 55.82 & 56.00 & 56.22 &22.38 & 22.40 & 23.10 \\ \midrule
Average & 11.76 & 12.18 & \textbf{12.95} & 58.43 & 58.43 & \textbf{58.62} & 27.78 & 27.67 & \textbf{28.57} \\ \bottomrule
\end{tabular}%
}
\caption{\llama~and \qwen~ICL results using the Dense Retrieval method for scoring with the \texttt{Script} vs.\ \texttt{IPA} vs.\ \texttt{Mixed} strategy for retrieval. Averaged over non-Latin script languages, our proposed \texttt{Mixed} retrieval strategy outperforms all other methods across all tasks.}
\label{tab:llama3-qwen2-dense-icl-results}
\end{table}

%% file: final_tables/appendix/gemma_mistral_icl.tex
\begin{table*}[htbp!]
\centering
\resizebox{\linewidth}{!}{%
\begin{tabular}{@{}cccccccccccccccc}
\toprule
\multirow{3}{*}{\begin{tabular}[c]{@{}c@{}}Gemma-\\  7B-Instruct\end{tabular}} & \multicolumn{5}{c}{\wiki~- BLEU ($\uparrow$)} & \multicolumn{5}{c}{\flores~- chrF ($\uparrow$)} & \multicolumn{5}{c}{\mlqa~- F1 ($\uparrow$)} \\ \cmidrule(lr){2-6} \cmidrule(lr){7-11} \cmidrule(lr){12-16}  
 & \multirow{2}{*}{0-shot} & \multirow{2}{*}{Random} & \multicolumn{3}{c}{BM25} & \multirow{2}{*}{0-shot} & \multirow{2}{*}{Random} & \multicolumn{3}{c}{BM25} & \multirow{2}{*}{0-shot} & \multirow{2}{*}{Random} & \multicolumn{3}{c}{BM25} \\ \cmidrule(lr){4-6} \cmidrule(lr){9-11} \cmidrule(l){14-16} 
 &  &  & Script & IPA & Mixed &  &  & Script & IPA & Mixed &  &  & Script & IPA & Mixed \\ \midrule
HIN &  11.52 & 40.92 & 42.25 & 42.28 & 42.83 
& 24.68 & 63.59 & 63.61 & 63.64 & 63.89
& 32.65 & 46.57 & 47.12 & 46.53 & 46.22\\
ARB & 2.52  & 25.92 & 27.41 & 26.12  & 28.03 
& 20.70 & 64.07 & 64.59 & 64.70 & 64.91 
& 20.99 & 29.98 & 30.22 & 30.81 & 32.35\\
ZHO & 1.29 & 6.49 & 8.42 & 9.43 & 9.28  
& 11.78 & 58.14 & 58.31 & 58.02 & 58.37  
& 10.28 & 12.76 & 13.68 & 12.64 & 13.17  \\
JPN & 1.84 & 9.49 & 12.72 & 12.04 & 13.12  
& 17.82 & 54.95 & 56.00 & 56.81 & 56.56  
& 21.46 & 24.20 & 23.24 & 24.34 & 25.15 \\ \midrule
Average &  4.29 & 20.71 & 22.70 & 22.47 & \cellcolor{darkspringgreen!20} \textbf{23.32} 
& 18.75 & 60.19 & 60.63 & 60.79 &  \cellcolor{darkspringgreen!20} \textbf{60.93}
&  21.35 & 28.38 & 28.57 & 28.58 & \cellcolor{darkspringgreen!20} \textbf{29.22} \\ \midrule
\multicolumn{1}{l}{} &  &  &  &  &  &  &  &  &  &  &  & \multicolumn{1}{l}{} &  &  &  \\ \midrule
\multirow{3}{*}{\begin{tabular}[c]{@{}c@{}}Mistral-\\  7B-Instruct\end{tabular}} & \multicolumn{5}{c}{\wiki~- BLEU ($\uparrow$)} & \multicolumn{5}{c}{\flores~- chrF ($\uparrow$)} & \multicolumn{5}{c}{\mlqa~- F1 ($\uparrow$)} \\ \cmidrule(lr){2-6} \cmidrule(lr){7-11} \cmidrule(lr){12-16}  
 & \multirow{2}{*}{0-shot} & \multirow{2}{*}{Random} & \multicolumn{3}{c}{BM25} & \multirow{2}{*}{0-shot} & \multirow{2}{*}{Random} & \multicolumn{3}{c}{BM25} & \multirow{2}{*}{0-shot} & \multirow{2}{*}{Random} & \multicolumn{3}{c}{BM25} \\ \cmidrule(lr){4-6} \cmidrule(lr){9-11} \cmidrule(l){14-16} 
 &  &  & Script & IPA & Mixed &  &  & Script & IPA & Mixed &  &  & Script & IPA & Mixed \\ \midrule
HIN & 4.63 & 40.65 & 42.25 & 42.12 & 42.77 
& 44.28 & 43.66 & 43.81 & 43.77 & 44.24 
& 16.46 & 36.74 & 37.29 & 36.51 & 39.46 \\
ARB & 1.47 & 25.24 & 26.24 & 24.91 & 27.00 
& 50.01 & 50.02 & 51.02 & 50.51 & 51.08 
& 16.09 & 28.66 & 29.81 & 29.15 & 30.42  \\
ZHO &  1.15 &  6.97 & 8.70 & 9.52 & 9.93  
&  51.53 & 53.01 & 53.52 & 53.12 & 53.54
& 5.11 & 11.15 & 13.58 & 12.06 & 13.05 \\
JPN & 1.73 &  9.61 & 11.44 & 11.55 & 12.18 
&  49.10 & 49.91 & 49.97 & 50.92 & 50.27
& 10.33 & 20.87 & 21.78 & 22.75 & 21.97  \\ \midrule
Average & 2.25 & 20.62 & 22.16 & 22.03 & \cellcolor{darkspringgreen!20} \textbf{22.97}
&  48.73 & 49.15 & 49.58 & 49.58 & \cellcolor{darkspringgreen!20} \textbf{49.78}
& 12.00 & 24.36 & 25.62 & 25.12  & \cellcolor{darkspringgreen!20} \textbf{26.23} \\ \bottomrule
\end{tabular}%
}
\caption{Gemma-7B-Instruct and Mistral-7B-Instruct ICL results using BM25 for scoring and the \texttt{Script} vs.\ \texttt{IPA} vs.\ \texttt{Mixed} strategy for retrieval. Averaged over non-Latin script languages, our proposed \texttt{Mixed} retrieval strategy outperforms all other methods on all tasks.}
\label{tab:gemma-mistral-icl-results}
\end{table*}

%% file: final_tables/main_result/gpt4-mixtral-icl-results-by-lang.tex
\begin{table*}[htbp!]
\centering
\resizebox{\linewidth}{!}{%
\begin{tabular}{@{}lccccccccccccccc@{}}
\toprule
\multicolumn{1}{c}{} & \multicolumn{5}{c}{\wiki~- BLEU ($\uparrow$)} & \multicolumn{5}{c}{\flores~- chrF ($\uparrow$)} & \multicolumn{5}{c}{\mlqa~- F1 ($\uparrow$)} \\ \cmidrule(lr){2-6} \cmidrule(lr){7-11} \cmidrule(lr){12-16} 
\multicolumn{1}{c}{} &  &  & \multicolumn{3}{c}{BM25} &  &  & \multicolumn{3}{c}{BM25} &  &  & \multicolumn{3}{c}{BM25} \\ \cmidrule(lr){4-6} \cmidrule(lr){9-11} \cmidrule(l){14-16} 
\multicolumn{1}{c}{\multirow{-3}{*}{GPT-4}} & \multirow{-2}{*}{0-shot} & \multirow{-2}{*}{Random} & Script & IPA & Mixed & \multirow{-2}{*}{0-shot} & \multirow{-2}{*}{Random} & Script & IPA & Mixed & \multirow{-2}{*}{0-shot} & \multirow{-2}{*}{Random} & Script & IPA & Mixed \\ \midrule
HIN & 12.32 & 22.14 & 47.04 & 38.38 & 30.53 & 77.16 & 72.53 & 76.09 & 73.49 & 97.30 & 65.77 & 59.15 & 59.15 & 61.11 & 58.16 \\
ARB & 20.91 & 28.11 & 25.96 & 8.85 & 27.16 & 85.22 & 70.58 & 69.49 & 81.41 & 72.78 & 51.85 & 49.62 & 50.75 & 48.48 & 50.75 \\
ZHO & 3.75 & 10.68 & 13.13 & 6.27 & 10.68 & 64.72 & 68.50 & 64.60 & 71.42 & 66.78 & 30.51 & 27.59 & 36.07 & 36.07 & 38.71 \\
JPN & 27.09 & 10.68 & 23.00 & 19.81 & 25.45 & 67.44 & 62.34 & 69.86 & 66.47 & 64.92 & 48.48 & 55.07 & 50.75 & 54.01 & 56.12 \\ \midrule
Average & 16.02 & 17.90 & \textbf{27.28} & 18.33 & 23.46 & 73.64 & 68.49 & 70.01 & 73.20 & \textbf{75.45} & 49.15 & 47.86 & 49.18 & 49.92 & \textbf{50.93} \\ \midrule
 & \multicolumn{3}{c}{} & \multicolumn{3}{c}{} & \multicolumn{3}{c}{} &  &  &  &  &  &  \\ \midrule
\multicolumn{1}{c}{} & \multicolumn{5}{c}{\wiki~- BLEU ($\uparrow$)} & \multicolumn{5}{c}{\flores ~- chrF ($\uparrow$)} & \multicolumn{5}{c}{\mlqa~- F1 ($\uparrow$)} \\ \cmidrule(lr){2-6} \cmidrule(lr){7-11} \cmidrule(lr){12-16} 
\multicolumn{1}{c}{} &  &  & \multicolumn{3}{c}{BM25} &  &  & \multicolumn{3}{c}{BM25} &  &  & \multicolumn{3}{c}{BM25} \\ \cmidrule(lr){4-6} \cmidrule(lr){9-11} \cmidrule(l){14-16} 
\multicolumn{1}{c}{\multirow{-3}{*}{\begin{tabular}[c]{@{}c@{}}Mixtral \\ 8x22B\end{tabular}}} & \multirow{-2}{*}{0-shot} & \multirow{-2}{*}{Random} & Script & IPA & Mixed & \multirow{-2}{*}{0-shot} & \multirow{-2}{*}{Random} & Script & IPA & Mixed & \multirow{-2}{*}{0-shot} & \multirow{-2}{*}{Random} & Script & IPA & Mixed \\ \midrule
HIN & 1.85 & 49.06 & 10.49 & 47.06 & 15.78 & 64.55 & 58.45 & 60.48 & 64.31 & 60.76 & 44.96 & 49.62 & 51.85 & 46.15 & 43.75 \\
ARB & 0.72 & 1.32 & 45.46 & 24.00 & 22.74 & 75.57 & 62.34 & 70.51 & 75.50 & 79.36 & 30.51 & 40.00 & 38.71 & 24.56 & 29.06 \\
ZHO & 4.46 & 0.81 & \textbf{4.37} & 2.86 & 9.86 & \textbf{0.00} & 58.68 & 64.25 & 62.24 & 73.59 & 29.06 & 26.09 & 31.93 & 26.09 & 31.93 \\
JPN & 1.65 & 5.16 & 23.00 & 1.21 & 2.73 & 62.68 & 29.23 & 60.03 & 62.65 & 64.47 & 29.06 & 46.15 & 44.96 & 37.40 & 41.27 \\ \midrule
Average & 2.17 & 14.09 & \textbf{20.83} & 18.78 & 12.78 & 50.70 & 52.18 & 63.82 & 66.17 & \textbf{69.55} & 33.40 & 40.47 & \textbf{41.86} & 33.55 & 36.50 \\ \bottomrule
\end{tabular}%
}
\caption{GPT-4 and Mixtral-8x22B-Instruct ICL results with various retrieval methods using \texttt{Script} vs.\ \texttt{IPA} vs.\ \texttt{Mixed} for retrieval (except 0-shot, all other columns represent results with 3-shot prompting). Prompts are task-agnostic for early exploration.}
\label{tab:gpt4-mixtral-icl-results-by-lang}
\end{table*}

%% file: final_tables/appendix/gpt4-mixtral-dense-results.tex
\begin{table}[htbp!]
\centering
\resizebox{\columnwidth}{!}{%
\begin{tabular}{@{}lccccccccc@{}}
\toprule
\multicolumn{1}{c}{\multirow{2}{*}{GPT-4}} & \multicolumn{3}{c}{\wiki} & \multicolumn{3}{c}{\flores} & \multicolumn{3}{c}{\mlqa} \\ \cmidrule(lr){2-4} \cmidrule(lr){5-7} \cmidrule(lr){8-10}  
\multicolumn{1}{c}{} & Script & IPA & Mixed & Script & IPA & Mixed & Script & IPA & Mixed \\ \midrule
HIN & 30.21 & 33.66 & 35.57 & 73.97 & 73.61 & 89.91 & 62.07 & 63.95 & 57.14 \\
ARB & 25.98 & 20.49 & 26.99 & 87.05 & 77.17 & 78.52 & 49.62 & 52.94 & 48.48 \\
ZHO & 8.91 & 9.43 & 12.19 & 65.50 & 72.82 & 68.50 & 30.51 & 36.07 & 37.40 \\
JPN & 26.99 & 10.68 & 12.70 & 68.69 & 73.34 & 68.83 & 51.85 & 51.85 & 54.01 \\ \midrule
Average & \textbf{23.02} & 18.57 & 21.86 & 73.80 & 74.24 & \textbf{76.44} & 48.51 & \textbf{51.20} & 49.26 \\ \midrule
 & \multicolumn{1}{l}{} & \multicolumn{1}{l}{} & \multicolumn{1}{l}{} & \multicolumn{1}{l}{} & \multicolumn{1}{l}{} & \multicolumn{1}{l}{} &  &  &  \\ \midrule
\multicolumn{1}{c}{\multirow{2}{*}{\begin{tabular}[c]{@{}c@{}}Mixtral \\ 8x22B\end{tabular}}} & \multicolumn{3}{c}{\wiki} & \multicolumn{3}{c}{\flores} & \multicolumn{3}{c}{\mlqa} \\ \cmidrule(lr){2-4} \cmidrule(lr){5-7} \cmidrule(lr){8-10}  
\multicolumn{1}{c}{} & Script & IPA & Mixed & Script & IPA & Mixed & Script & IPA & Mixed \\ \midrule
HIN & 37.21 & 41.17 & 9.60 & 60.29 & 65.12 & 78.12 & 42.52 & 49.62 & 42.52 \\
ARB & 29.17 & 6.96 & 39.13 & 76.91 & 66.67 & 76.20 & 42.52 & 38.71 & 38.71 \\
ZHO & 1.30 & 0.84 & 15.14 & 58.05 & 52.79 & 68.44 & 36.07 & 30.51 & 31.93 \\
JPN & 2.57 & 13.60 & 2.77 & 60.73 & 60.82 & 64.52 & 51.85 & 43.75 & 46.15 \\ \midrule
Average & \textbf{17.56} & 15.64 & 16.66 & 64.00 & 61.35 & \textbf{71.82} & \textbf{43.24} & 40.65 & 39.83 \\ \bottomrule
\end{tabular}%
}
\caption{GPT-4 and Mixtral-8x22B ICL results using the Dense Retrieval method using \texttt{Script} vs.\ \texttt{IPA} vs.\ \texttt{Mixed} strategy for retrieval. Prompts are task-agnostic for early exploration.}
\label{tab:gpt4-mixtral-dense-icl-results}
\end{table}

%% file: final_tables/ablations/abbr_tab/detailed_mixing_variations.tex
% \usepackage{multirow}
\begin{table}[htbp!]
\resizebox{\columnwidth}{!}{%
\begin{tabular}{@{}cccccccc@{}}
\toprule
 & \multirow{-1}{*}{ \begin{tabular}[c]{@{}c@{}} Divide- \\ Conquer \end{tabular}} 
 & \multirow{2}{*}{Harmonic} & \multicolumn{3}{c}{Split-Half} & \multirow{2}{*}{Concat} & \multirow{-1}{*}{ \begin{tabular}[c]{@{}c@{}} Mix \\ (Ours) \end{tabular}}  \\ \cmidrule(lr){4-6} 
 &  &  & IPA+Script & Script+IPA & Shuffle &  &  \\ \midrule
\wiki & 21.34 & 21.61 & 22.15 & 22.14 & 22.02 & 21.62 & \textbf{22.28} \\
\flores & 57.52 & 57.33 & 57.13 & 57.53 & 57.50 & 57.62 & \textbf{57.84} \\
\mlqa & 28.94 & 28.87 & 28.48 & 28.17 & 28.55 & 28.72 & \textbf{30.97} \\  \bottomrule
\end{tabular}%
}

\caption{Comparing 4 different mixing strategies under 6-shot settings. Even number of shot is required for comparison with \texttt{Split-Half} approach. \llama~is leveraged as the base LLM.}

\label{tab:abbrv_abl_mixing_variations}
\end{table}

%% file: final_tables/ablations/abbr_tab/icl_sft_table.tex
\begin{table}[htbp!]
\centering
\resizebox{\columnwidth}{!}{%
\begin{tabular}{@{}ccccccc@{}}
\toprule
\multirow{2}{*}{Task} & \multirow{2}{*}{0-shot} & \multicolumn{2}{c}{M2Lingual SFT} & \multicolumn{3}{c}{ICL} \\ \cmidrule(lr){3-4} \cmidrule(lr){5-7} 
 &  & Script data & Script + IPA data & 3-shot & 6-shot & 10-shot \\ \midrule
\wiki & 2.86 & 8.86 & 6.46 & 19.85 & 22.28 & \textbf{23.16} \\
\flores & 49.67 & 18.15 & 19.74 & 57.44 & 57.84 & \textbf{58.03} \\
\mlqa & 20.84 & 23.60 & 24.74 & 30.75 & 30.97 & \textbf{31.40} \\ \bottomrule
\end{tabular}%
}
\caption{Comparison between BM25 \mixicl~and two variations of SFT models on M2Lingual dataset: (1) Script-only data and (2) Script+IPA data. ICL approach is evaluated under 3-shot, 6-shot and 10-shot settings and SFT methods are under 0-shot evaluation. \llama~is leveraged as the base LLM.}
\label{tab:abbrv_icl_sft}
\end{table}

%% file: final_tables/appendix/promptopt_quality_mlqa_compare.tex
\begin{table}[htbp!]
\resizebox{\columnwidth}{!}{%
\begin{tabular}{c|c|c|cc}
& \ayamlqa & \multicolumn{3}{c}{\orimlqa} \\ \hline
& Ours & Ours & XLM FT & mBERT FT \\ \hline 
HIN & 49.30 & 63.46 & 34.40 & 50.20 \\
ARB & 30.49 & 55.74 & 42.50 & 52.30 \\
ZHO & 14.38 & 43.10 & 40.50 & 59.60 \\ \hline
\textbf{Average} & \textbf{31.39} & \textbf{54.10} & \textbf{39.13} & \textbf{54.03}
\end{tabular}%
}
\caption{Performance evaluation comparison between \ayamlqa~ and \orimlqa~with \llama~and reported baseline of XLM and mBERT PLM. } 
\label{tab:appendix_aya_ori_mlqa_compare}
\end{table}

%% file: final_tables/appendix/llama3-latin-performance.tex
\begin{table*}[t]
\centering
\resizebox{\linewidth}{!}{%
\begin{tabular}{@{}cccccccccccccccc@{}}
\toprule
\multirow{3}{*}{\begin{tabular}[c]{@{}c@{}}Llama3-\\  8B-Instruct\end{tabular}} & \multicolumn{5}{c}{\wiki - BLEU ($\uparrow$)} & \multicolumn{5}{c}{\flores - chrF ($\uparrow$)} & \multicolumn{5}{c}{\mlqa- F1 ($\uparrow$)} \\ \cmidrule(lr){2-6} \cmidrule(lr){7-11} \cmidrule(lr){12-16}  
 & \multirow{2}{*}{0-shot} & \multirow{2}{*}{Random} & \multicolumn{3}{c}{BM25} & \multirow{2}{*}{0-shot} & \multirow{2}{*}{Random} & \multicolumn{3}{c}{BM25} & \multirow{2}{*}{0-shot} & \multirow{2}{*}{Random} & \multicolumn{3}{c}{BM25} \\ \cmidrule(lr){4-6} \cmidrule(lr){9-11} \cmidrule(l){14-16} 
 &  &  & Script & IPA & Mixed &  &  & Script & IPA & Mixed &  &  & Script & IPA & Mixed \\ \midrule
DEU & 11.50 & 28.04 & 31.53 & 31.38 & 31.82 & 62.02 & 66.16 & 66.43 & 66.54 & 66.54 & 40.67 & 47.63 & 45.91 & 44.52 & 48.50 \\
FRA & 15.84 & 35.93 & 41.18 & 40.37 & 40.86 & 57.47 & 67.05 & 67.24 & 67.35 & 67.32 &12.26 & 12.54 & 13.11 & 12.69 & 14.09 \\
SPA & 12.08 & 39.23 & 42.33 & 42.40 & 42.38 & 51.57 & 60.10 & 60.38 & 60.43 & 60.42 & 24.70 & 27.28 & 27.45 & 27.99 & 27.33 \\
POR & 18.25 & 30.61 & 34.92 & 35.83 & 35.57 & 60.45 & 70.18 & 70.30 & 70.52 & 70.66 & 40.52 & 44.96 & 46.74 & 52.62 & 48.56 \\ \midrule
\textbf{Average} & \textbf{14.42} & \textbf{33.45} & \textbf{37.49} & \textbf{37.50} & \textbf{37.66} & \textbf{57.88} & \textbf{65.87} & \textbf{66.09} & \textbf{66.21} & \textbf{66.24} & \textbf{29.54} & \textbf{33.10} & \textbf{33.30} & \textbf{34.46} & \textbf{34.62} \\ \bottomrule
\end{tabular}%
}
\caption{\llama~ICL results using the BM25 retrieval method using \script~vs \ipa~vs \mixed~retrieval strategy for Latin-based languages (DEU, FRA, SPA, POR).}
\label{tab:llama3-latin-results}
\end{table*}